\documentclass[12pt,letterpaper]{article}
\usepackage{style/dsc180reportstyle} 
\usepackage{float}

\title{Investigation into In-Context
Learning Capabilities of
Transformers}

\author{
  Rushil Chandrupatla \\
  {\tt ruchandrupatla@ucsd.edu} \\
  \And
  Leo Bangayan \\
  {\tt lbangayan@ucsd.edu} \\
  \And
  Sebastian Leng \\
  {\tt sjleng@ucsd.edu} \\
}

\begin{document}
\maketitle


\begin{abstract}
Transformers have demonstrated a strong ability for in-context learning (ICL), enabling models to solve previously unseen tasks using only example input–output pairs provided at inference time. While prior theoretical work has established conditions under which transformers can perform linear classification in-context, the empirical scaling behavior governing when this mechanism succeeds remains insufficiently characterized.

In this paper, we conduct a systematic empirical study of in-context learning for Gaussian-mixture binary classification tasks. Building on the theoretical framework of Frei and Vardi (2024), we analyze how in-context test accuracy depends on three fundamental factors: the input dimension, the number of in-context examples, and the number of pre-training tasks. Using a controlled synthetic setup and a linear in-context classifier formulation, we isolate the geometric conditions under which models successfully infer task structure from context alone.

We additionally investigate the emergence of benign overfitting, where models memorize noisy in-context labels while still achieving strong generalization performance on clean test data. Through extensive sweeps across dimensionality, sequence length, task diversity, and signal-to-noise regimes, we identify the parameter regions in which this phenomenon arises and characterize how it depends on data geometry and training exposure.

Our results provide a comprehensive empirical map of scaling behavior in in-context classification, highlighting the critical role of dimensionality, signal strength, and contextual information in determining when in-context learning succeeds and when it fails.
\end{abstract}

\begin{center}
Code: \url{https://github.com/Shou-Yue/DSC180a-ICL-A11/tree/rushil}
\end{center}

\maketoc
\clearpage


\section{Introduction}
\subsection{Introduction}
Transformers are a type of artificial intelligence model that uses a mechanism known as attention to determine the relationships between different components of input (which can be text, images, or data). Transformers serve as the backbone for state of the art machine learning models we see today such as ChatGPT. 

When these models are first created, they have no base knowledge. In order for them to be usable, we must train them on some data relevant to the task we want the model to be able to complete. For example, if we want a model to filter through loan applications and choose who to apply or deny, we would train the model on past loan application data, along with the final result. The model will then find patterns between our characteristic data (for example, this may include gender, income, credit score, etc.) and the final result. This is known as unsupervised learning and concludes our training process. Then, we can feed the model new loan applications, and it will apply its learned pattern to determine if the application should be approved or denied very quickly. 

While this seems very efficient, the caveat is that training can take a very long time, especially when we want the model to be able to do larger tasks. For example, the training for GPT-3 took around 3-4 continuous months. This leads into the mechanisms of in-context learning. In-context learning is the idea that a model can be given examples of a task that it hasn’t been trained on and thus hasn’t seen before. Then, without training the model anymore, we can directly use it for these new tasks, saving all that training time and compute power. Thus, understanding how this mechanism works and how we can best leverage it can lead to massive cutdowns in model training time, allowing faster breakthroughs in the machine learning field.

\subsection{Literature Review}
We have read and worked with several papers to build an understanding of in-context learning in transformers for both regression and classification tasks. Below is a summary of relevant papers.

\paragraph{Case Study, Function Classes} \cite{garg2023} examines more closely the relationship between the training data and the tasks on which in-context learning succeeds. To achieve this, the paper attempts training transformers on data derived by function classes (ie. linear functions, two-layer neural networks, decision trees). The transformer is then given input and output pairs of the same task but a new function at test time and is asked to predict the output for some new input. Through empirical testing, the paper shows that transformers can be pre-trained to learn new linear functions at test time solely based on in-context examples, with performance comparable to the optimal least squares estimator. This idea is then extended to the more complex function classes and is shown to match performance of optimal task-based learning algorithms, Furthermore, this paper also shows that transformers can perform in-context learning successfully even when the original training data and new inference prompts differ (ie. model was trained on 2D data but asked to predict on 5D data) or when the underlying data distribution between the in-context examples and query input differ.

\paragraph{Linear Classification} \cite{frei2024} extends the work of recent papers from linear regression to classification tasks. This paper looks into the environment needed for successful in-context learning; more specifically, the number of pre-training tasks and in-context examples that are needed for the transformer to generalize well at test time. They found that generalization is a result of the complex interworkings between the number of tasks, ICL examples, noise, and dimensions. While they were unable to zero in on a specific formula to find the number of such tasks and examples, they did find that generalization occurs as a result of transformers implementing a max-margin type classifier in its forward pass. The paper also exhibits the idea of benign overfitting: under specific circumstances, transformers are able to generalize well at test time even when noise is present in the original dataset, such as classes swapped for binary classification tasks. The paper also found that cluster separation affects ICL training and testing accuracy greatly. At lower separation, overfitting occurs. At higher separation, test accuracy improves drastically. Between these areas is where benign overfitting occurs.

\subsection{Problem Statement}
We will proceed with a case study, in which we will train different transformers for various linear classification tasks, then evaluate their ability to generalize to new examples through in-context learning and the circumstances in which benign overfitting occurs. From this, we hope to gain a better understanding of both the nuances of in-context learning: where does it work best, when can it be relied upon and maybe when should it not be relied on, as well as more insight into the idea of and when benign overfitting occurs and how it can be leveraged to cut down on training time and compute power for machine learning models.  

We will first define a precise technical problem, then delve into precise research questions we hope to answer through this project.

We consider \textbf{linear binary classification tasks} where data for each task $\tau$ is generated via the class-conditional Gaussian mixture model:
\[
\mu_\tau \sim \mathrm{Unif}(R \cdot S^{d-1}), 
\qquad 
y_{\tau,i} \in \{-1, 1\},
\qquad
x_{\tau,i} = y_{\tau,i}\mu_\tau + z_{\tau,i}, 
\qquad 
z_{\tau,i} \sim \mathcal{N}(0, I_d).
\]

Each task consists of a small sequence of in-context examples:
\[
\{(x_{\tau,1}, y_{\tau,1}), \ldots, (x_{\tau,N}, y_{\tau,N})\},
\]
followed by a query point $x_{\tau,N+1}$ whose label must be predicted.

The goal is to analyze how a trained transformer maps the \emph{in-context sequence} to a prediction rule for $x_{\tau,N+1}$ \emph{without updating parameters}.

We seek to understand:
\begin{itemize}
    \item Under what regimes of dimension $d$,
    \item number of tasks (pre-training samples) $B$,
    \item sequence length $N$,
    \item and signal-to-noise ratio $R_\tau$,
\end{itemize}
transformers perform near-optimal in-context learning.

We extend this task to 3 broad research questions that we aim to answer.

\subsection{Key Research Questions}
\textbf{RQ1}: For transformers trained on Gaussian mixture tasks, how does in-context test accuracy scale with dimension d, number of tasks B, and sequence length N?

\textbf{RQ2}: Under what conditions does the model exhibit benign overfitting, i.e., memorize noisy in-context labels while still achieving high test accuracy?

\textbf{RQ3}: Are the phenomena described theoretically for linear attention transformers preserved when using full transformer architectures (decoder-only, encoder-only)? How do the results change when full transformer architectures are used?

Prior work has shown that in-context learning can be effective for binary classification tasks, but only analyzing a restricted linear attention architecture. These works also primarily focus on theory, but no comprehensive empirical explorations have been done on this topic to date. No existing work systematically tests what we have set out to explore in our research questions: ICL behavior across different transformer types and models, how benign overfitting manifests empirically, and how changing variables (dimension, batch size, sequence length, signal strength, and noise) affect generalization. These are the gaps that our work will cover. We seek to answer these questions through comprehensive empirical testing, changing one input at a time and examining its effects. We hope that this will lead to a better understanding of how in-context learning and benign overfitting can best be leveraged to reduce training time and compute power needed for large-scale machine learning models.

\section{Methods}
\subsection{Research Question 1}
\subsubsection{Model Architecture}

Following the theoretical formulation of in-context classification studied by Frei and Vardi (2024), we employ a linear in-context learning classifier designed to isolate the geometric mechanism of task inference without introducing nonlinear attention components.

The model consists of a single learnable matrix $W \in \mathbb{R}^{d \times d}$. Given a sequence $E$ containing $N$ labeled context examples and a query input $x_{N+1}$, the model first computes the label-weighted empirical mean of the context:
\[
\hat{\mu} = \frac{1}{N}\sum_{i=1}^{N} y_i x_i
\]
where labels are represented as $y_i \in \{-1,+1\}$.

The prediction for the query is then given by
\[
\hat{y}(E;W) = \hat{\mu}^{\top} W x_{N+1}.
\]

Thus, the learned matrix $W$ acts as a preconditioning transformation that reshapes the inferred task vector prior to computing the classification score. This formulation mirrors the convex linear-transformer parameterization analyzed in prior theoretical work while allowing us to directly study scaling behavior under controlled synthetic task distributions.

\subsubsection{Training Procedure}

We train the model using stochastic gradient descent on the logistic loss applied to the query prediction. At each optimization step, a batch of $B$ independent synthetic classification tasks is generated. Each task contains $N$ context examples and a single query example.

Unless otherwise specified, all experiments use the following training configuration:

\begin{itemize}
    \item \textbf{Optimization:} Stochastic gradient descent with fixed learning rate $\eta = 0.01$ and no weight decay.
    \item \textbf{Initialization:} The weight matrix $W$ is initialized to zero.
    \item \textbf{Training duration:} Each run is trained for at most 1000 optimization steps.
    \item \textbf{Evaluation frequency:} Validation metrics are computed every 10 training steps using newly sampled tasks with identical structural parameters but independent randomness.
    \item \textbf{Reproducibility:} Training and validation batches are generated using distinct deterministic seeds to ensure both reproducibility and statistical independence.
\end{itemize}

The loss is computed only on the query example for each task; context predictions are used exclusively for evaluation.

\subsubsection{Evaluation Metrics}

To evaluate model performance, we report three accuracy measures averaged over independently generated validation tasks:

\begin{enumerate}
    \item \textbf{Training Accuracy.} This metric measures the proportion of correctly classified query examples on the batch used for the gradient update.
    
    \item \textbf{Validation Accuracy (In-Context Test Accuracy).} This metric measures the proportion of correctly classified query examples on newly sampled tasks. 
    
    \item \textbf{In-Context Accuracy.} This metric measures the model’s ability to correctly classify the labeled examples within the context sequence itself when applying the inferred task representation. High in-context accuracy indicates stronger memorization of the provided examples.
\end{enumerate}

Both validation accuracy and in-context accuracy measure aspects of the in-context learning performance. The main difference between the two is that validation accuracy is measured on truly unseen tasks, while in-context accuracy is measured on tasks that were used to set the boundary (and thus theoretically should be be easier to classify) but still unseen by the model during training. For each configuration, we record the full learning trajectory as well as summary statistics including the best validation accuracy achieved and the training step at which near-optimal performance is first reached.

\subsection{Experiments}

We conduct a series of controlled experiments designed to measure how in-context test accuracy scales with input dimension, context size, and number of tasks.

\paragraph{Dimensionality Analysis}
To understand how ambient feature dimension affects in-context learning, we vary the input dimension
\[
d \in \{50,100,200,500,1000\}.
\]
For these experiments, we fix the context size $N=20$ and the number of tasks per batch $B=1000$. We measure validation accuracy to determine how representational dimensionality influences task inference and classification performance.

\paragraph{Context Size Analysis}
To evaluate how the number of in-context examples affects learning, we vary the context length
\[
N \in \{5,10,20,40,80\}
\]
while holding $d=500$ and $B=1000$ fixed. This experiment tests how additional examples improve the model’s ability to infer the underlying task distribution.

\paragraph{Task Batch Size Analysis}
To study the role of task diversity during training, we vary the number of tasks per batch
\[
B \in \{50,100,250,500,1000,2000\}
\]
while fixing $d=500$ and $N=20$. This experiment examines whether exposure to more simultaneous tasks improves generalization performance.

\paragraph{Signal-to-Noise Ratio Regimes}
To control for how task separability scales with dimension, we evaluate two signal regimes:
\begin{itemize}
    \item A constant signal magnitude $R=6.45$
    \item A dimension-scaled signal magnitude $R=0.3\sqrt{d}$
\end{itemize}
These regimes allow us to distinguish whether performance trends arise from dimensional scaling itself or from implicit changes in signal-to-noise ratio.

\paragraph{Interaction Grids}
To capture interactions between scaling variables, we additionally evaluate two-dimensional grids over $(d,N)$ and $(B,N)$. These experiments allow us to observe whether the effect of additional context depends on dimensionality or task diversity.

\paragraph{Replication Strategy}
Each configuration is trained using three independent random seeds, and all reported results correspond to the mean and standard deviation across seeds.
\subsection{Research Question 2}

\subsubsection{Objective}

To investigate when benign overfitting occurs, we introduce controlled label noise into the in-context examples and analyze the relationship between:

\begin{itemize}
    \item In-context memorization accuracy
    \item Query test accuracy
    \item Signal-to-noise ratio
    \item Dimension and context size
\end{itemize}

We define \textbf{benign overfitting} as the regime in which the model achieves high in-context accuracy on noisy labels while maintaining high validation (clean test) accuracy. This indicates memorization of corrupted context labels without degradation in generalization.

\subsubsection{Noise Injection Procedure}

We modify the Gaussian mixture task generation as follows.

For each task $\tau$, we first generate clean labels:
\[
y_{\tau,i} \in \{-1, +1\}.
\]

We then independently flip each context label with probability $\epsilon$:
\[
y_{\tau,i}^{\text{noisy}} =
\begin{cases}
- y_{\tau,i}, & \text{with probability } \epsilon, \\
\;\; y_{\tau,i}, & \text{with probability } 1-\epsilon.
\end{cases}
\]

We evaluate noise levels
\[
\epsilon \in \{0, 0.05, 0.1, 0.2, 0.3, 0.4\}.
\]

Noise is injected only into context labels; query labels remain clean. This allows us to isolate memorization behavior from generalization performance.

\subsubsection{Evaluation Metrics}

We will be using the same evaluation metrics as those defined in Section 2.1.3. Evaluation metrics will be tracked every 50 steps.

\subsubsection{Experimental Sweeps}

To determine when benign overfitting emerges, we perform controlled parameter sweeps.

\paragraph{Noise Level Sweep.}

We fix:
\[
d = 500, \quad N = 20, \quad B = 1000,
\]
and evaluate both signal regimes (constant $R$ and dimension-scaled $R$).

We vary:
\[
\epsilon \in \{0, 0.05, 0.1, 0.2, 0.3, 0.4\}.
\]

This identifies the critical noise threshold beyond which generalization collapses.

\paragraph{Signal-to-Noise Interaction.}

We vary signal strength $R$ across low, medium, and high separation regimes while fixing $\epsilon = 0.2$. This tests whether benign overfitting occurs only in intermediate separation regimes, as suggested by prior theoretical work.

\paragraph{Dimensionality Interaction.}

We vary:
\[
d \in \{100, 200, 500, 1000\},
\]
while holding:
\[
N = 20, \quad B = 1000, \quad \epsilon = 0.2.
\]

This evaluates whether higher-dimensional regimes facilitate benign overfitting through effective overparameterization.

\paragraph{Context Size Interaction.}

We vary:
\[
N \in \{5, 10, 20, 40, 80\},
\]
with fixed:
\[
d = 500, \quad \epsilon = 0.2.
\]

This tests whether longer context sequences increase memorization capacity relative to signal strength.

\subsubsection{Regime Classification}

We classify observed behaviors into three categories:

\begin{enumerate}
    \item \textbf{Underfitting:} Low in-context accuracy and low validation accuracy.
    \item \textbf{Classical Overfitting:} High in-context accuracy and low validation accuracy.
    \item \textbf{Benign Overfitting:} High noisy in-context accuracy and high validation accuracy.
\end{enumerate}

We visualize these regimes using phase diagrams over $(d, \epsilon)$, $(R, \epsilon)$, and $(N, \epsilon)$.

\subsection{Research Question 3: Full Transformer Architectures}

To address RQ3 ("Are the phenomena described theoretically for linear attention transformers preserved when using full transformer architectures?"), we extended our analysis beyond the trained linear transformer to evaluate off-the-shelf, pre-trained full transformer architectures. We employed a two-pronged approach: probing open-weights models on regression tasks and testing commercial Large Language Models (LLMs) on classification tasks.

\subsubsection{Commercial Model Testing (GMM Classification)}
We developed a testing framework to evaluate commercial LLMs (specifically gpt-4o-mini) on the same Gaussian Mixture Model (GMM) binary classification tasks used in RQ1. Since these models expect text input rather than raw tensors, we implemented a serialization procedure:
\begin{itemize}
    \item \textbf{Data Formatting:} Each numerical feature vector $x \in \mathbb{R}^d$ was formatted as a comma-separated string of floating-point numbers truncated to four decimal places.
    \item \textbf{Prompt Structure:} We constructed few-shot prompts containing $N$ labeled context examples followed by a single unlabeled query point. The prompt explicitly instructed the model to act as a binary classification model and output only the predicted label (0 or 1).
    \item \textbf{Evaluation:} We queried the model API with varying configurations of dimension ($d$), context length ($N$), and signal strength ($R$). Predictions were parsed and compared against the ground truth labels generated by the GMM.
\end{itemize}

\subsubsection{Open-Weights Model Probing (Linear Regression)}
To test if full transformers naturally implement gradient descent (GD) algorithms as suggested by linear attention theory, we probed the \texttt{TinyLlama-1.1B} model on synthetic linear regression tasks.
\begin{itemize}
    \item \textbf{Task Setup:} We sampled random linear tasks $y = Wx$ where inputs $x \sim U(-\alpha, \alpha)$.
    \item \textbf{Baselines:} We compared the model's Mean Squared Error (MSE) against two theoretical baselines: one-step Gradient Descent (GD-1) initialized at zero, and a preconditioned variant (GD++).
    \item \textbf{OOD Testing:} We further evaluated the model on non-linear Sine wave tasks to test for "benign overfitting" or algorithmic breakdown on out-of-distribution data.
\end{itemize}

\section{Results}
\subsection{Research Question 1}
Notable graphs from the experiments outlined above are shown below.
\subsubsection{Dimension Experiments}
\begin{figure}[htbp]
  \centering
  \includegraphics[width=1.1\textwidth]{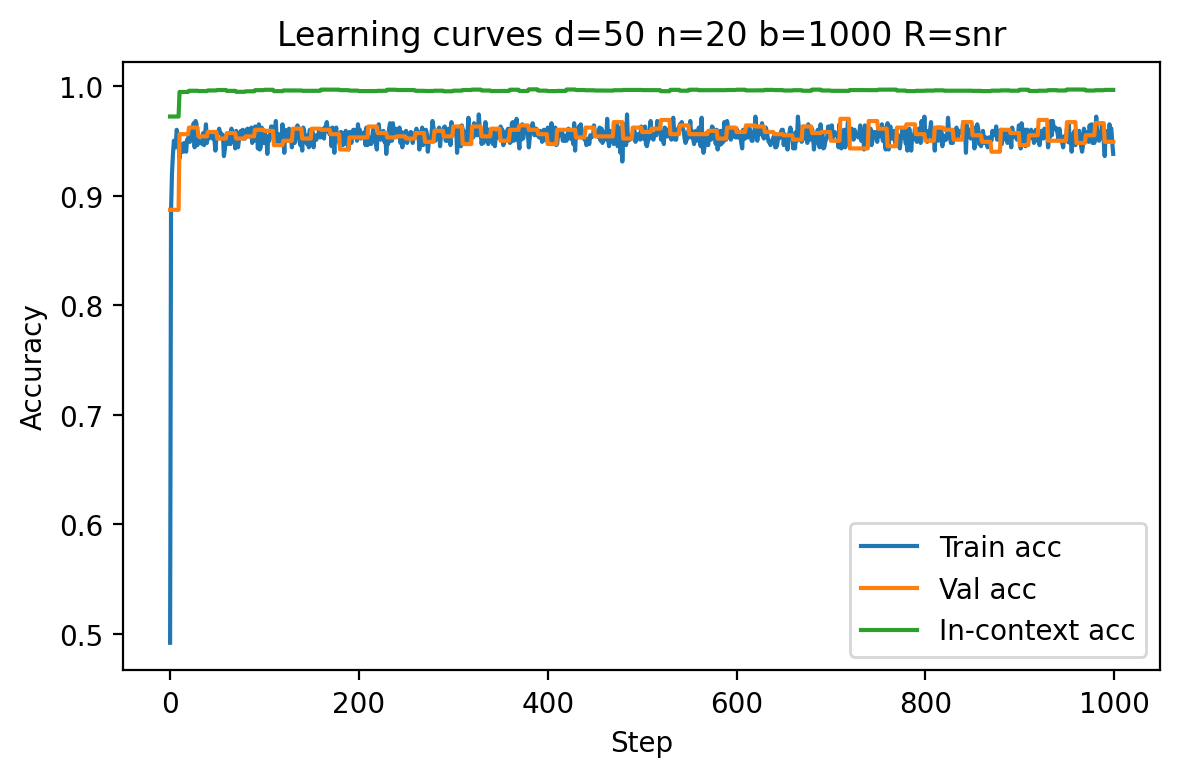}
  \caption{Model Performance (d=50, N=20, B=1000, R=snr)}
  \label{fig:high_dim_1}
\end{figure}
\begin{figure}[htbp]
  \centering
  \includegraphics[width=1.1\textwidth]{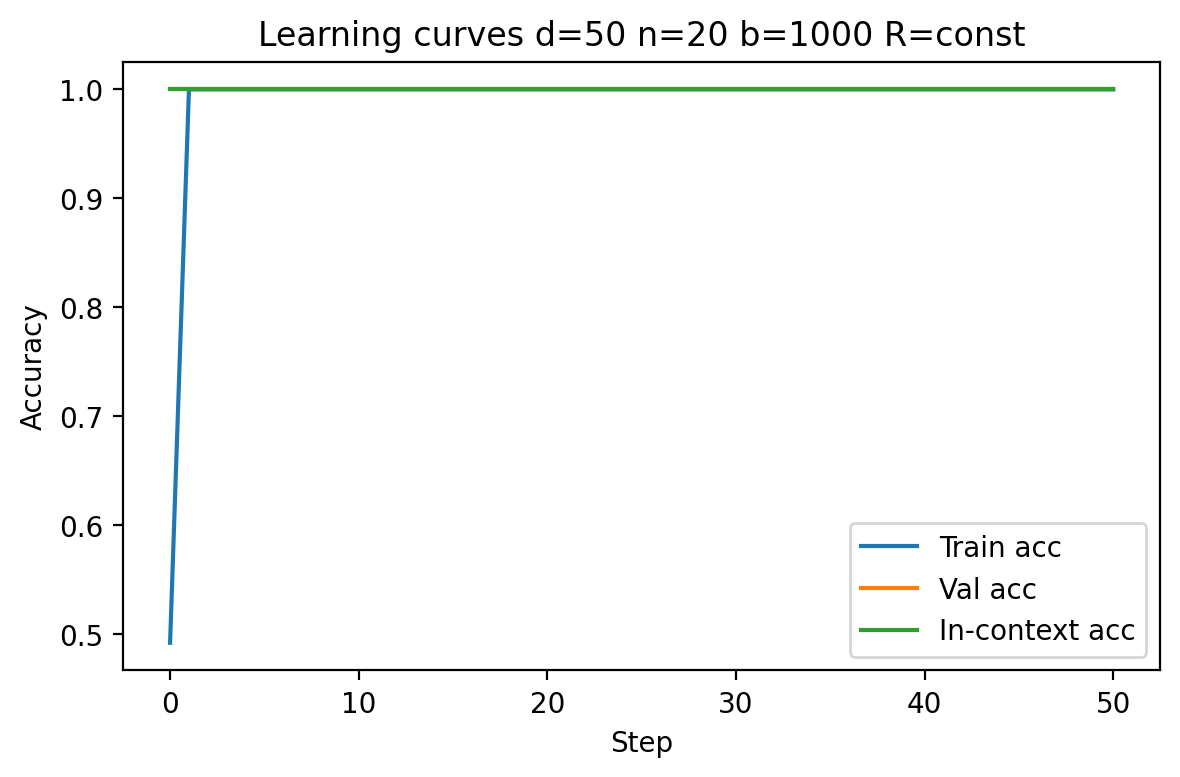}
  \caption{Model Performance (d=50, N=20, B=1000, R=const)}
  \label{fig:high_dim_1}
\end{figure}
\begin{figure}[htbp]
  \centering
  \includegraphics[width=1.1\textwidth]{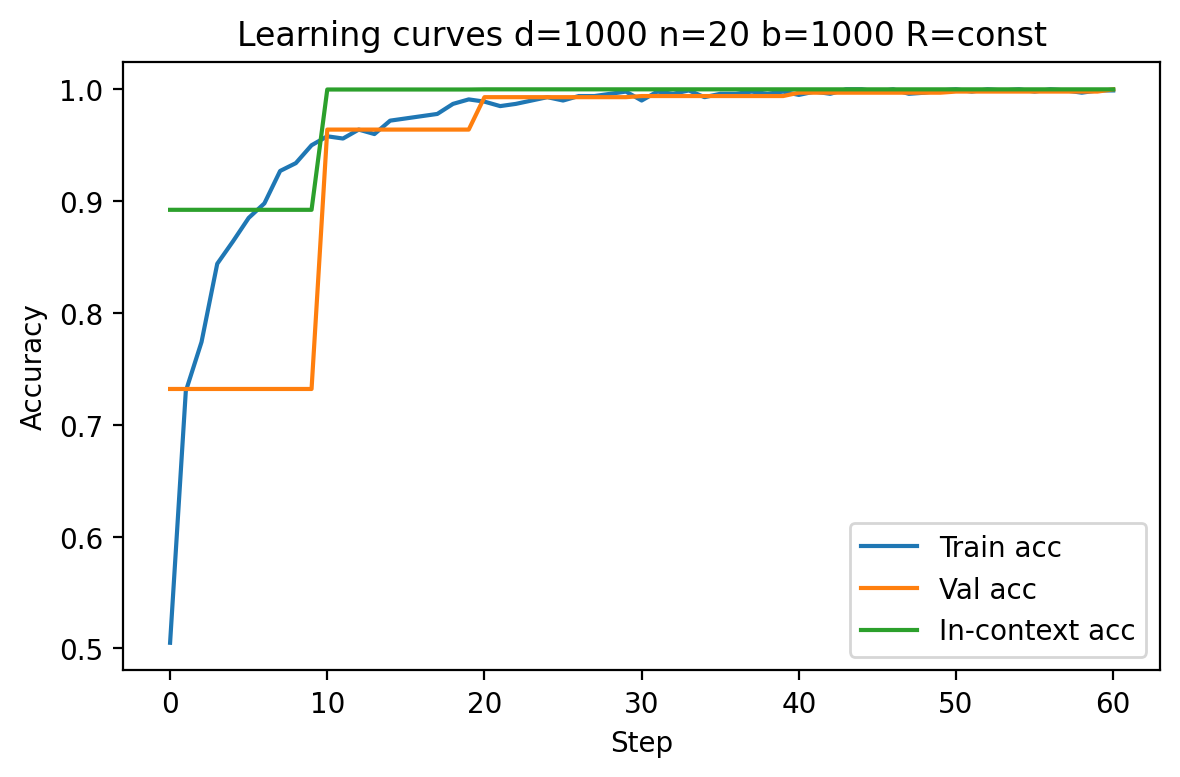}
  \caption{Model Performance (d=1000, N=20, B=1000, R=const)}
  \label{fig:high_dim_1}
\end{figure}
\begin{figure}[htbp]
  \centering
  \includegraphics[width=1.1\textwidth]{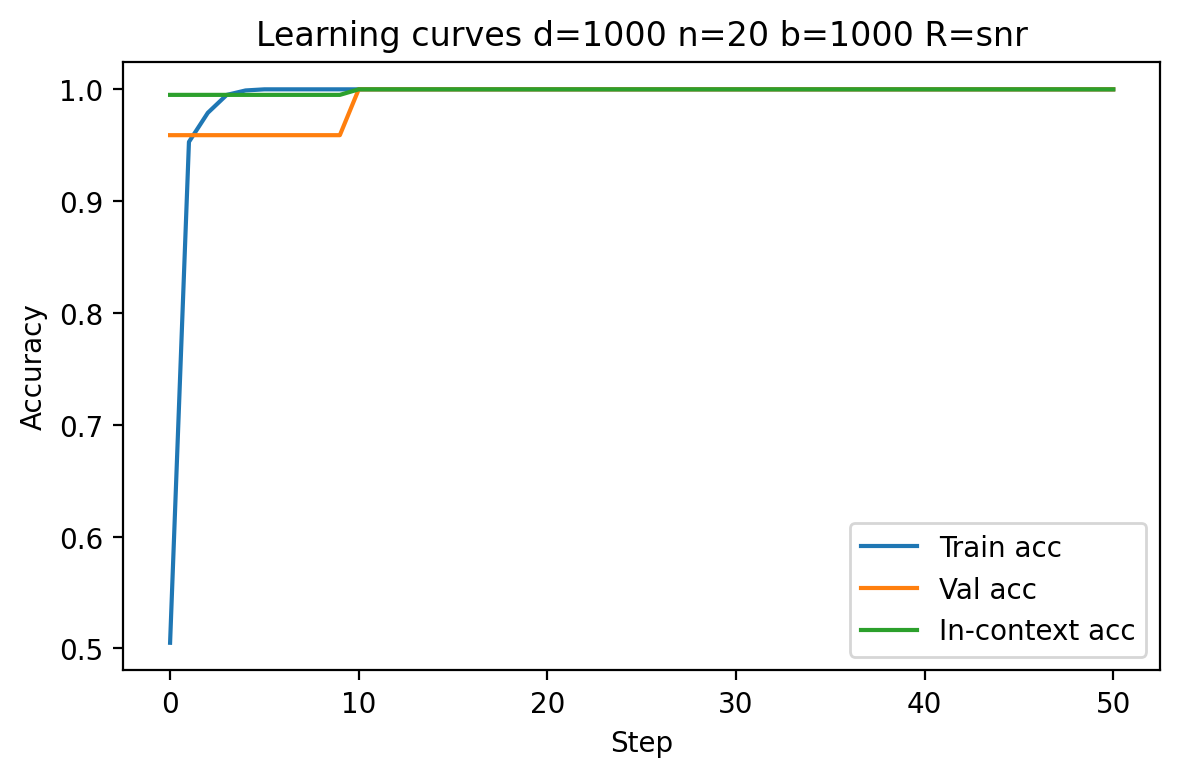}
  \caption{Model Performance (d=1000, N=20, B=1000, R=snr)}
  \label{fig:high_dim_1}
\end{figure}

\subsubsection{Sequence Length Experiments}
\begin{figure}[H]
  \centering
  \includegraphics[width=1.1\textwidth]{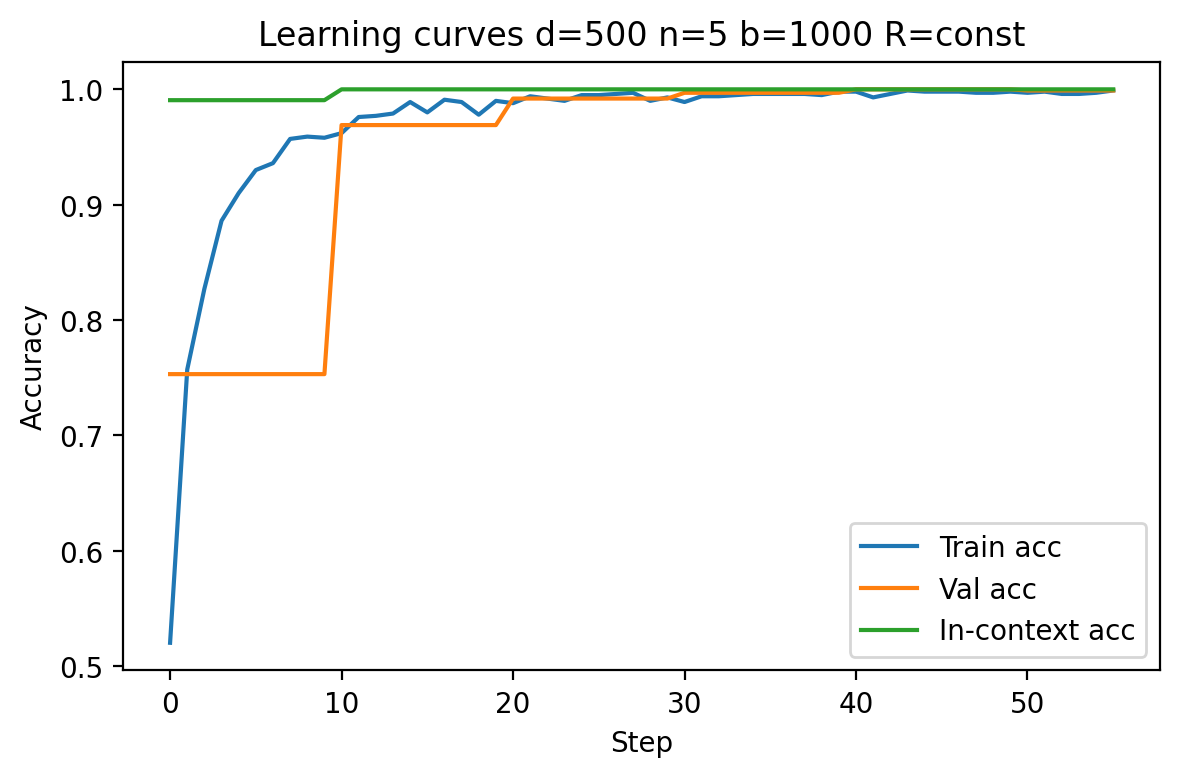}
  \caption{Model Performance (d=500, N=5, B=1000, R=const)}
  \label{fig:high_dim_1}
\end{figure}
\begin{figure}[H]
  \centering
  \includegraphics[width=1.1\textwidth]{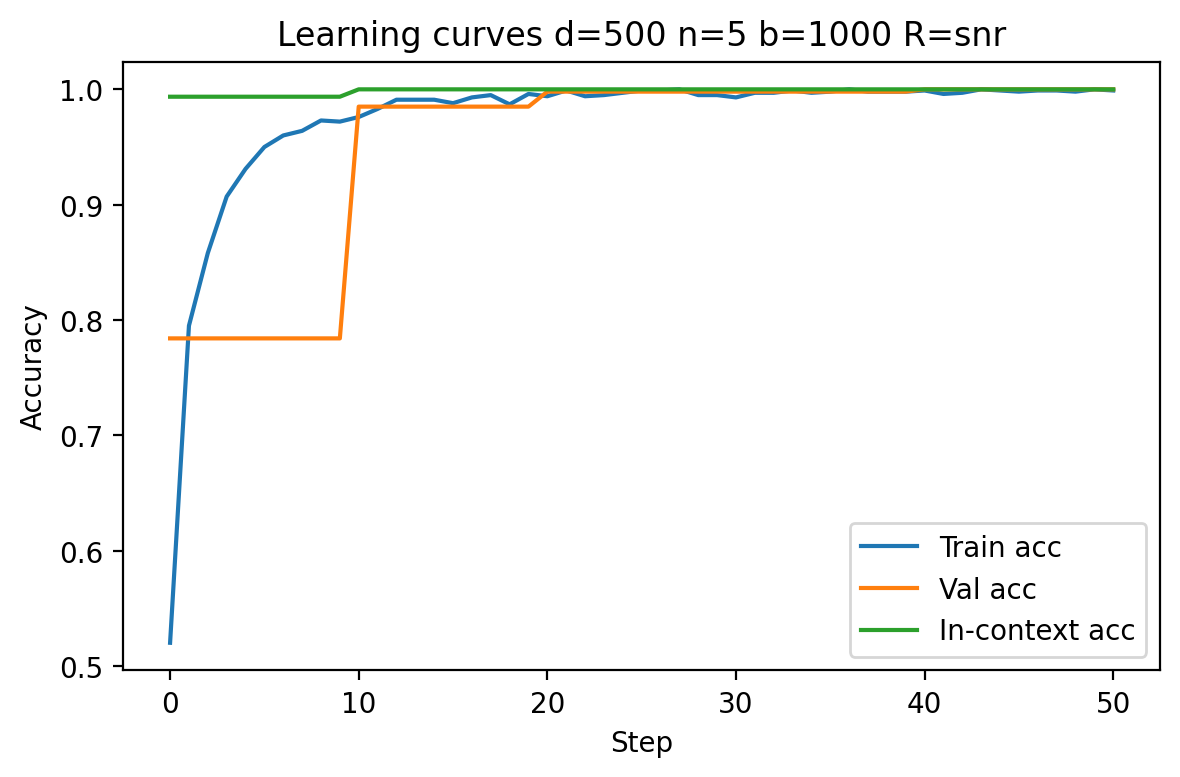}
  \caption{Model Performance (d=500, N=5, B=1000, R=snr)}
  \label{fig:high_dim_1}
\end{figure}
\begin{figure}[H]
  \centering
  \includegraphics[width=1.1\textwidth]{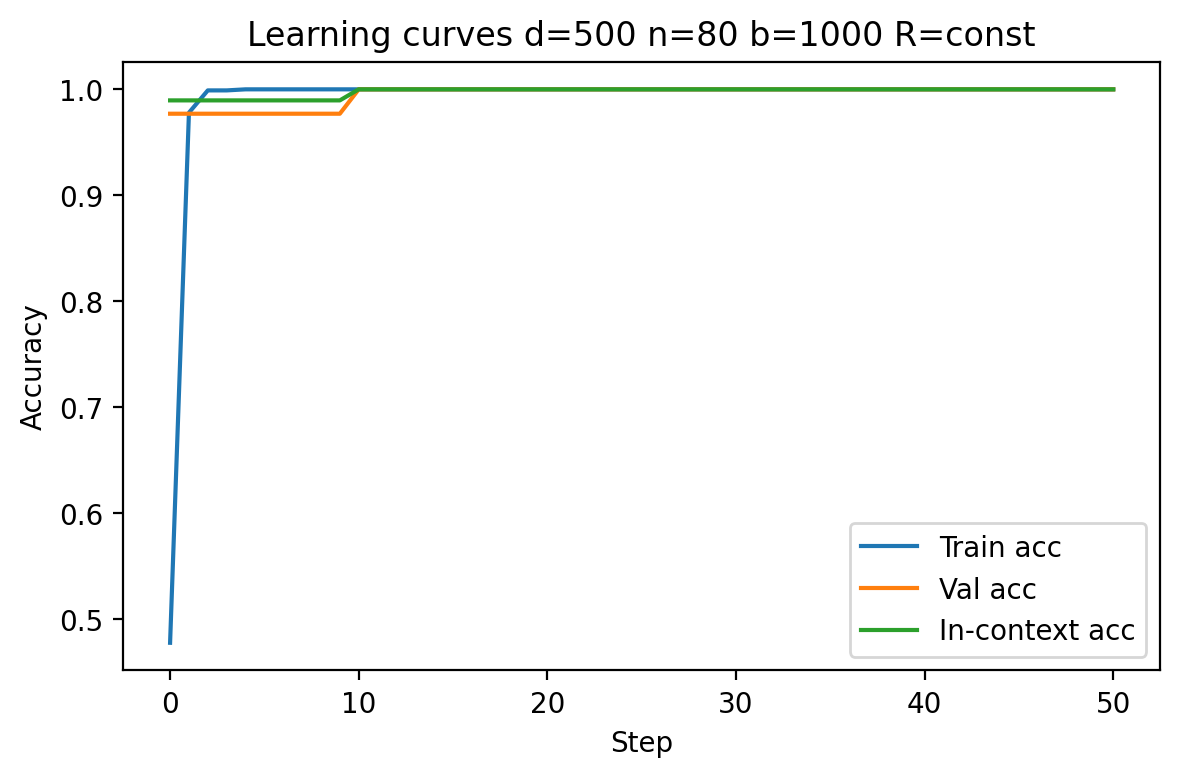}
  \caption{Model Performance (d=500, N=80, B=1000, R=const)}
  \label{fig:high_dim_1}
\end{figure}
\begin{figure}[H]
  \centering
  \includegraphics[width=1.1\textwidth]{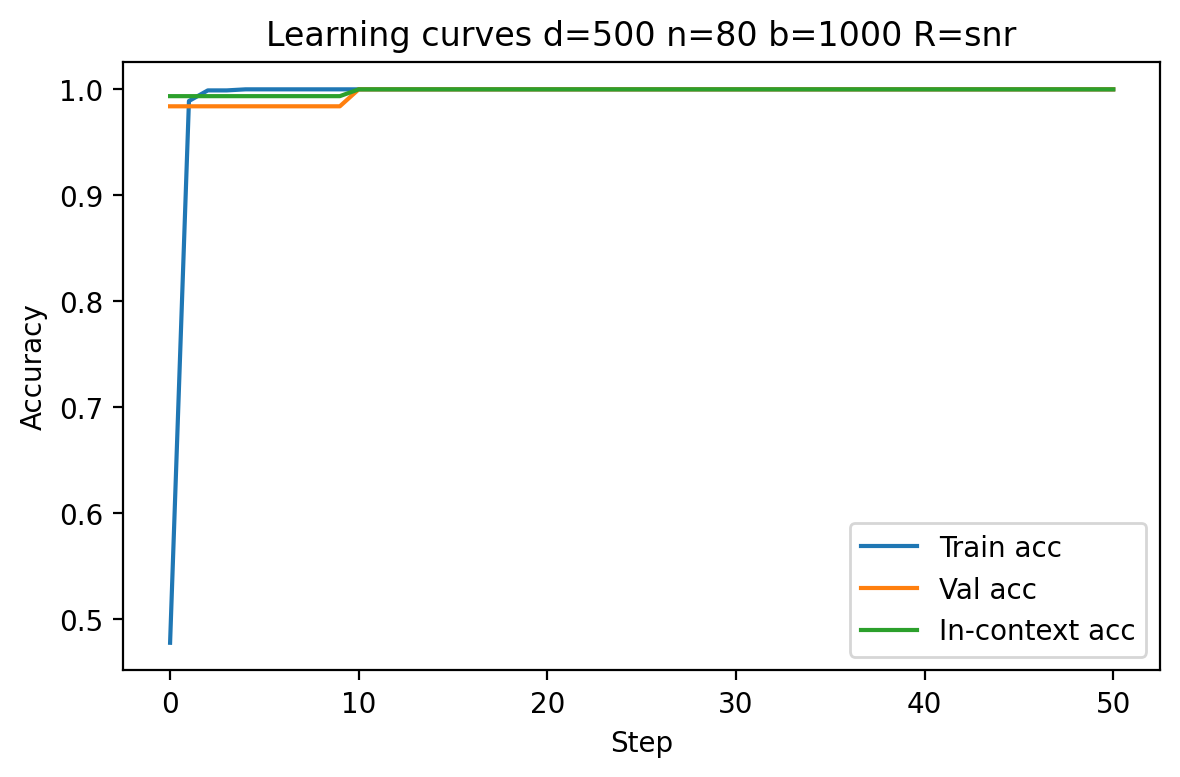}
  \caption{Model Performance (d=500, N=80, B=1000, R=snr)}
  \label{fig:high_dim_1}
\end{figure}

\subsubsection{Batch Size Experiments}
\begin{figure}[H]
  \centering
  \includegraphics[width=1.1\textwidth]{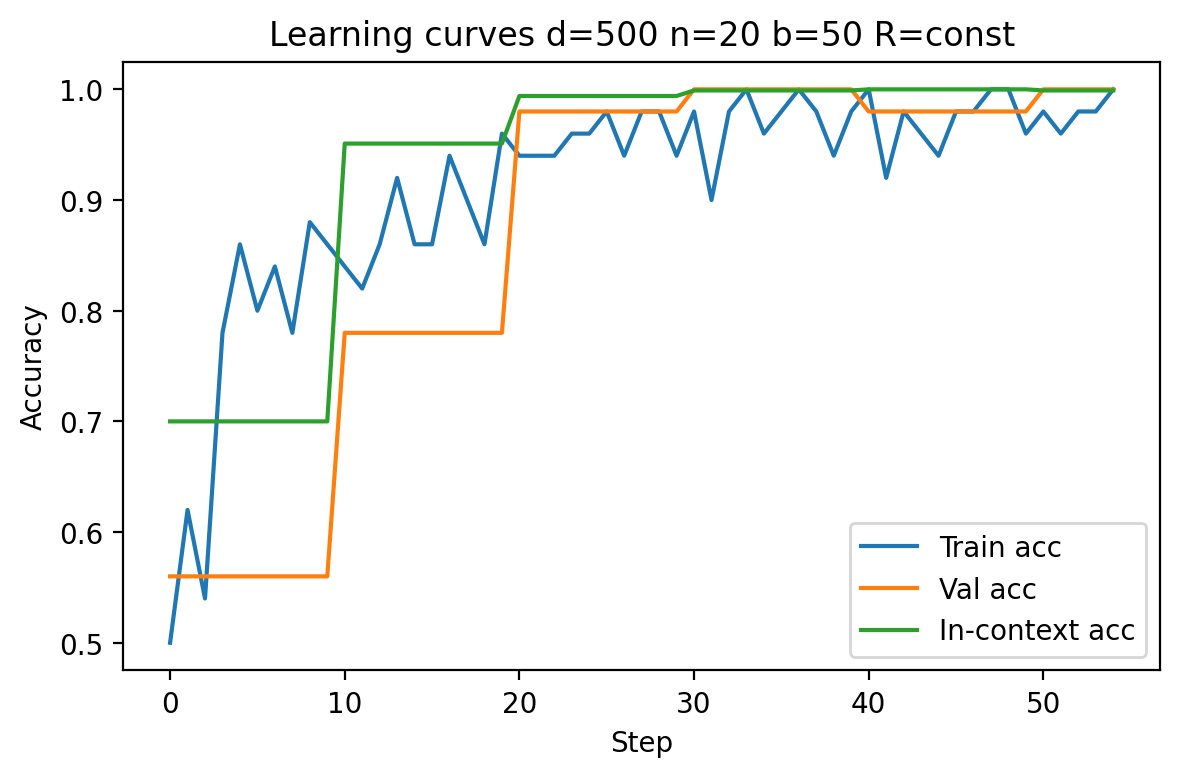}
  \caption{Model Performance (d=500, N=20, B=50, R=const)}
  \label{fig:high_dim_1}
\end{figure}
\begin{figure}[H]
  \centering
  \includegraphics[width=1.1\textwidth]{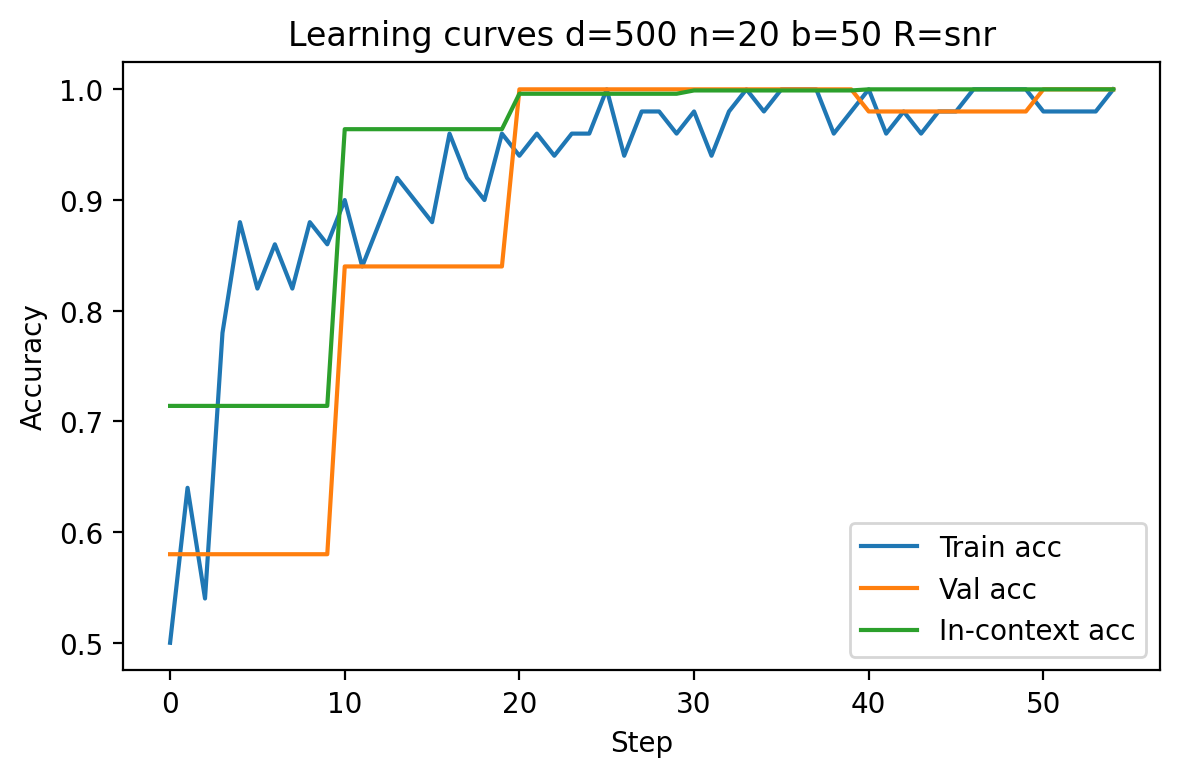}
  \caption{Model Performance (d=500, N=20, B=50, R=snr)}
  \label{fig:high_dim_1}
\end{figure}
\begin{figure}[H]
  \centering
  \includegraphics[width=1.1\textwidth]{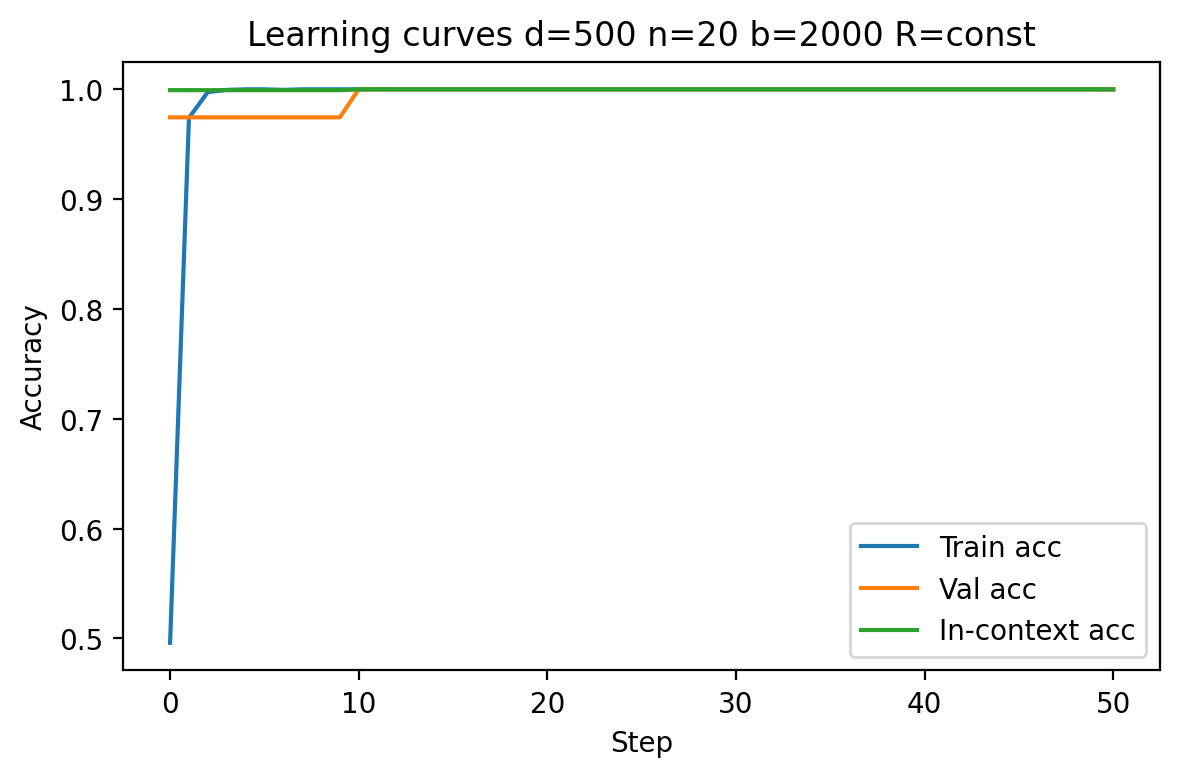}
  \caption{Model Performance (d=500, N=20, B=2000, R=const)}
  \label{fig:high_dim_1}
\end{figure}
\begin{figure}[H]
  \centering
  \includegraphics[width=1.1\textwidth]{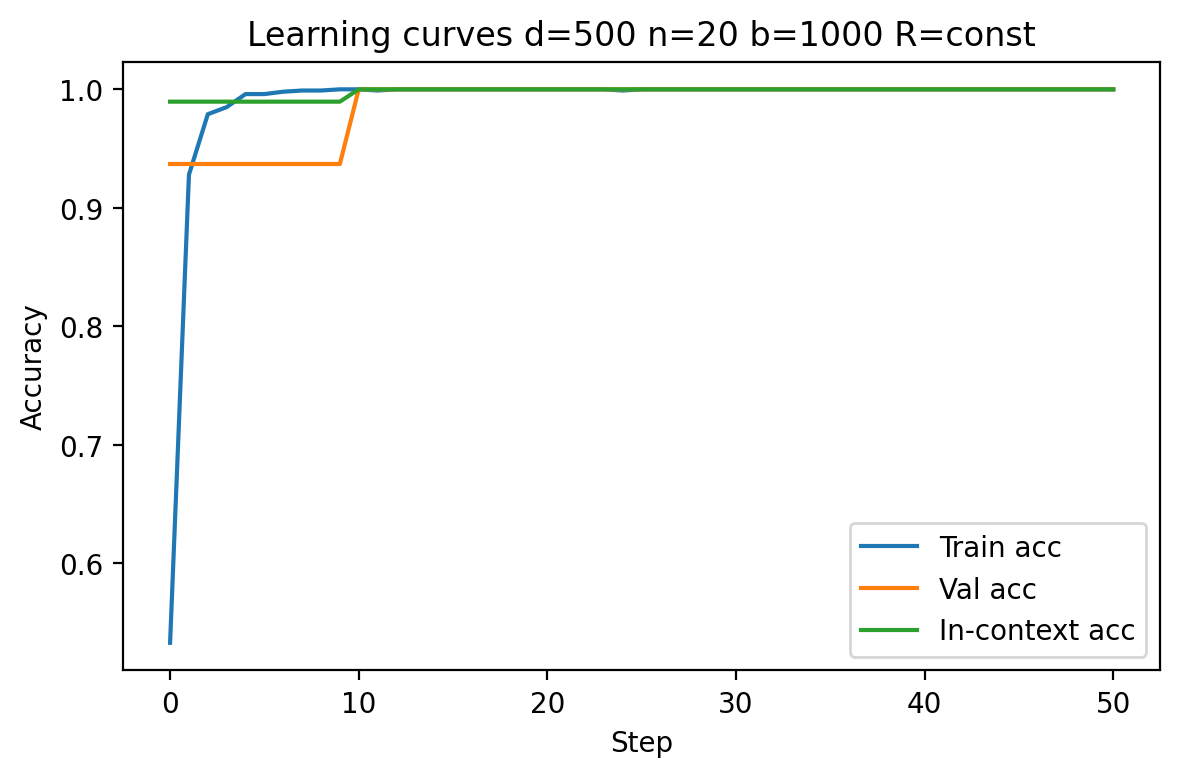}
  \caption{Model Performance (d=500, N=20, B=1000, R=const)}
  \label{fig:high_dim_1}
\end{figure}

\subsection{Research Question 2}
\subsubsection{Noise applied to Context Labels}
\begin{figure}[H]
  \centering
  \includegraphics[width=1.1\textwidth]{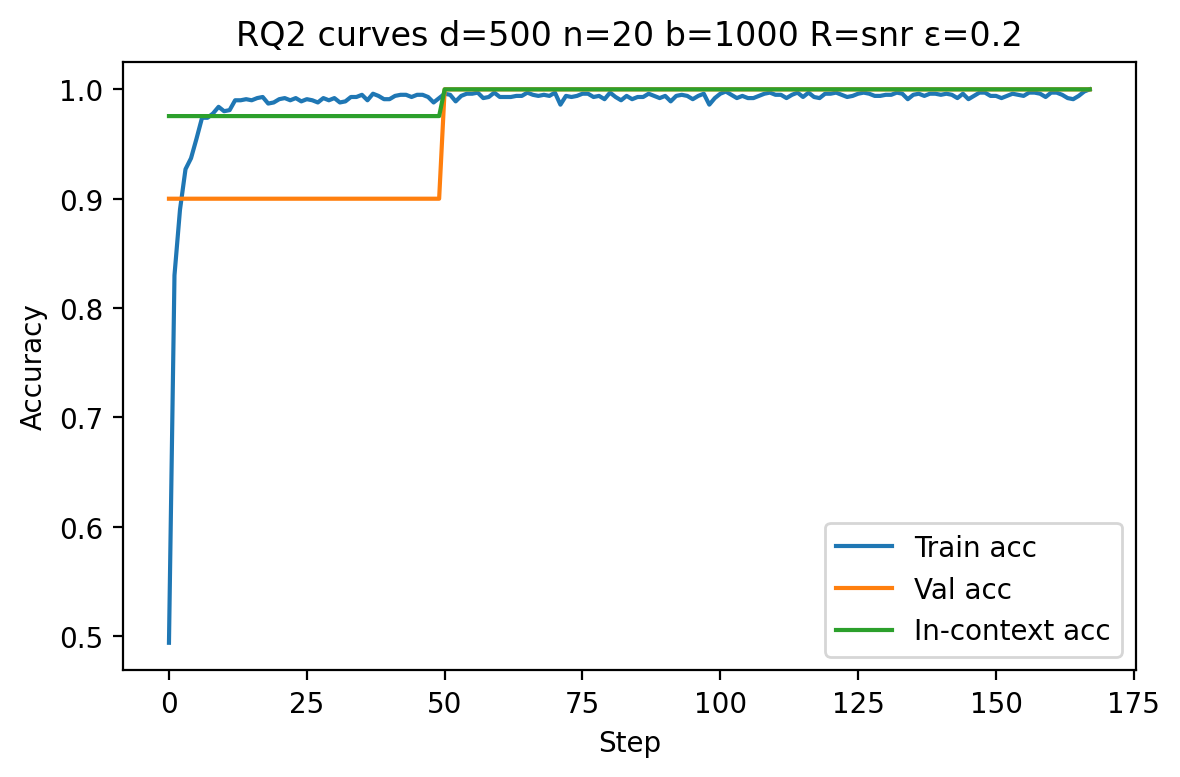}
  \caption{Model Performance (d=500, N=20, B=1000, R=snr, noise=0.20)}
  \label{fig:high_dim_1}
\end{figure}
\begin{figure}[H]
  \centering
  \includegraphics[width=1.1\textwidth]{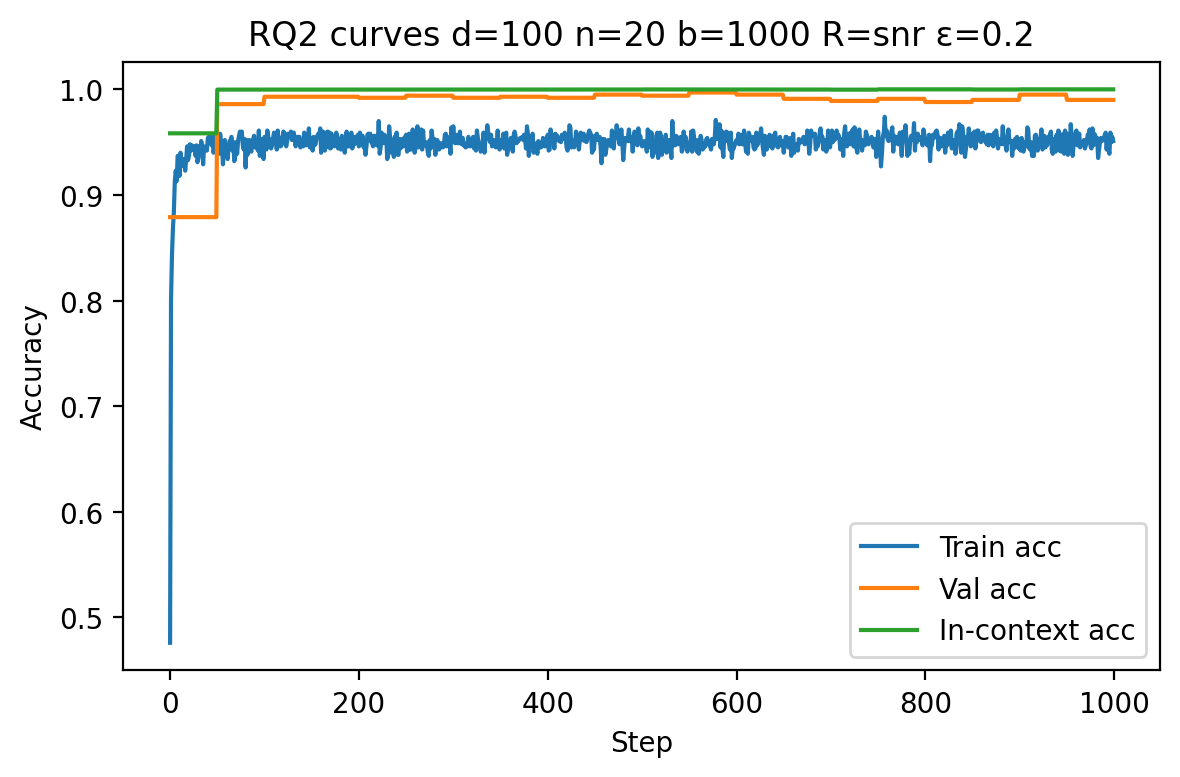}
  \caption{Model Performance (d=100, N=20, B=1000, R=snr, noise=0.20)}
  \label{fig:high_dim_1}
\end{figure}
\begin{figure}[H]
  \centering
  \includegraphics[width=1.1\textwidth]{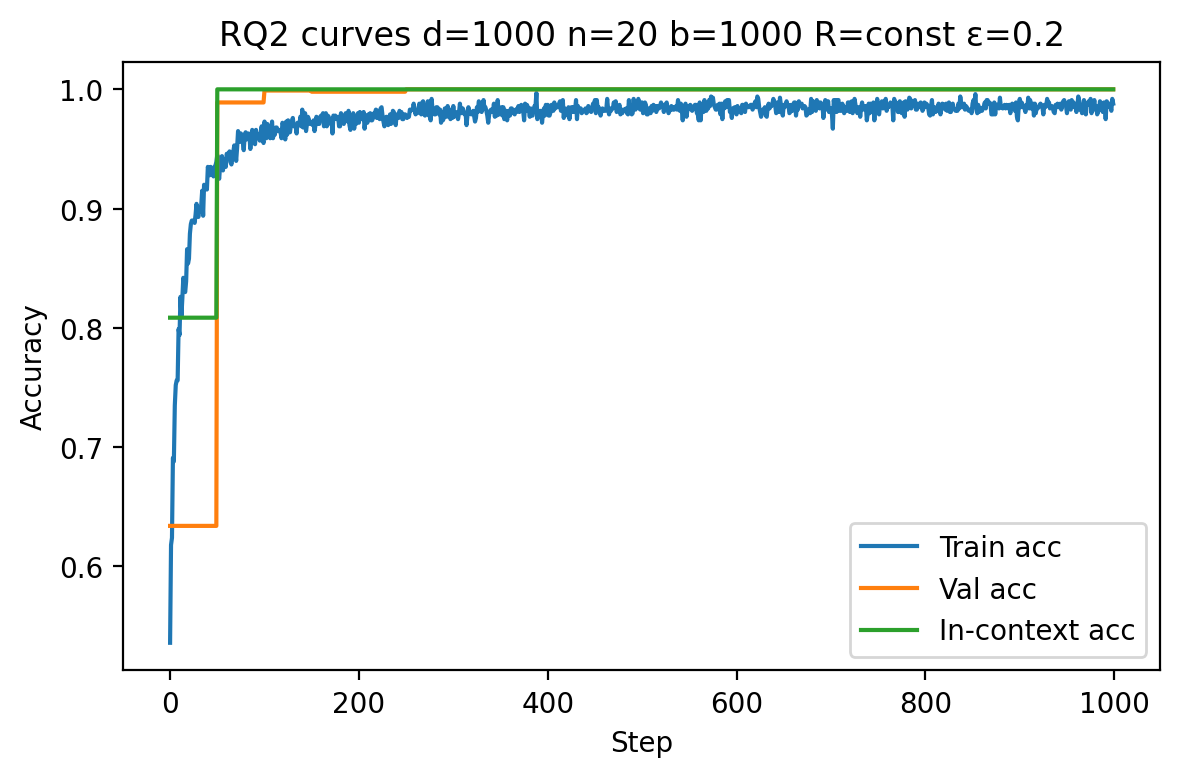}
  \caption{Model Performance (d=1000, N=20, B=1000, R=const, noise=0.20)}
  \label{fig:high_dim_1}
\end{figure}
\begin{figure}[H]
  \centering
  \includegraphics[width=1.1\textwidth]{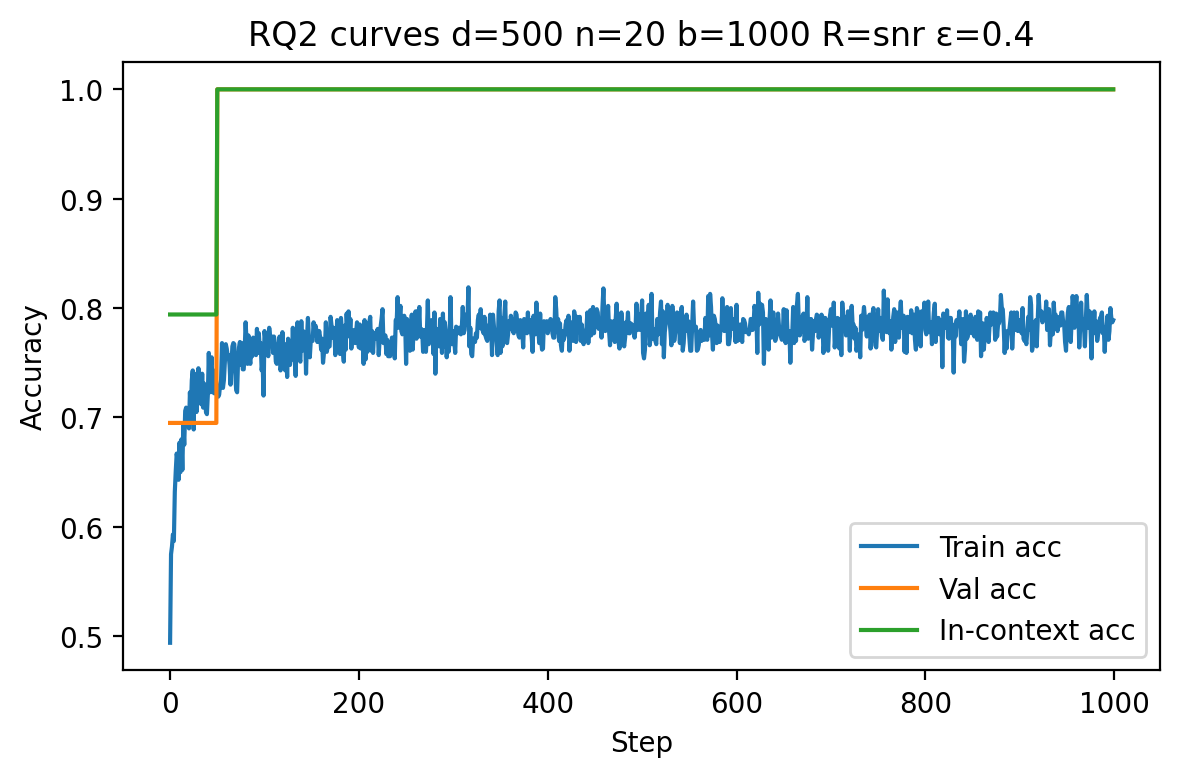}
  \caption{Model Performance (d=500, N=20, B=1000, R=snr, noise=0.40)}
  \label{fig:high_dim_1}
\end{figure}

\subsubsection{Noise applied to Context and Query Labels}

\begin{figure}[H]
  \centering
  \includegraphics[width=1.1\textwidth]{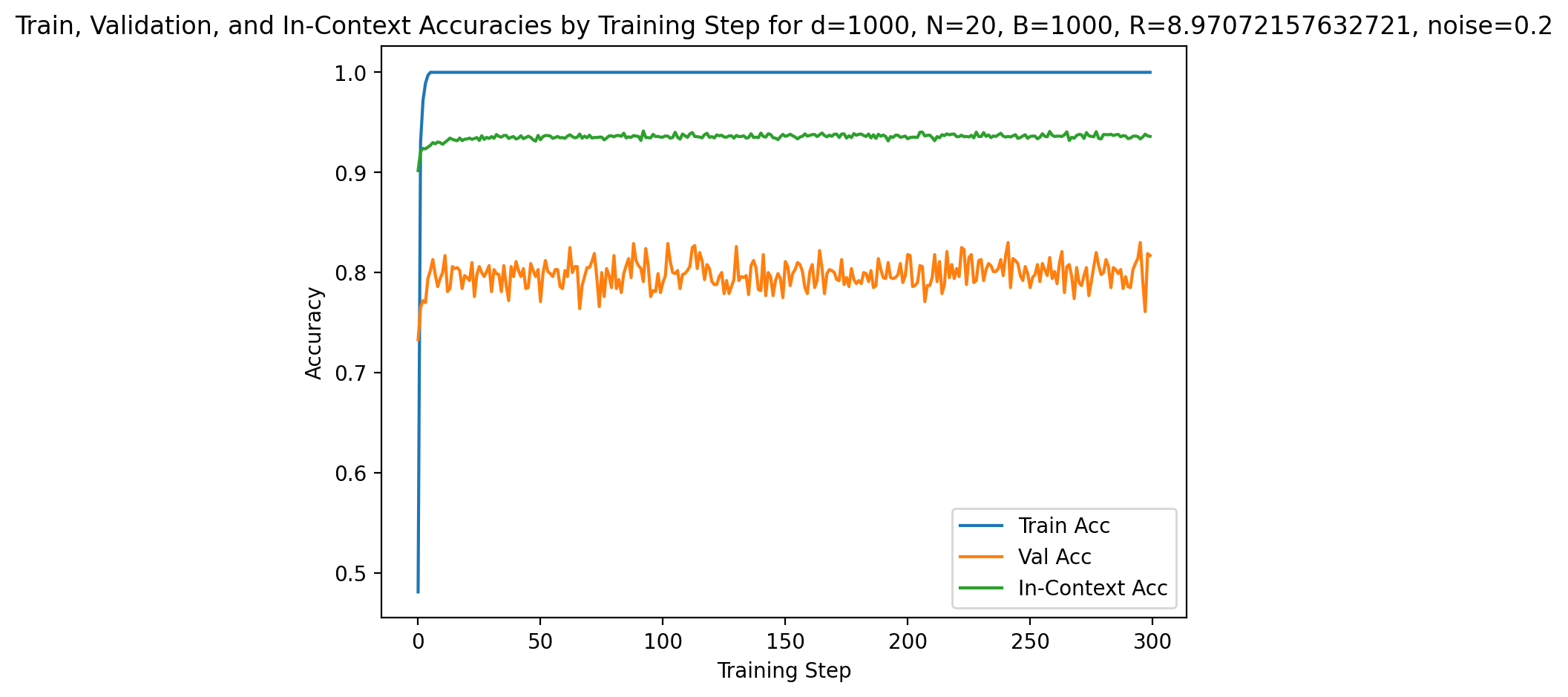}
  \caption{Model Performance (Benign Overfitting Case)}
  \label{fig:high_dim_1}
\end{figure}
\begin{figure}[H]
  \centering
  \includegraphics[width=1.1\textwidth]{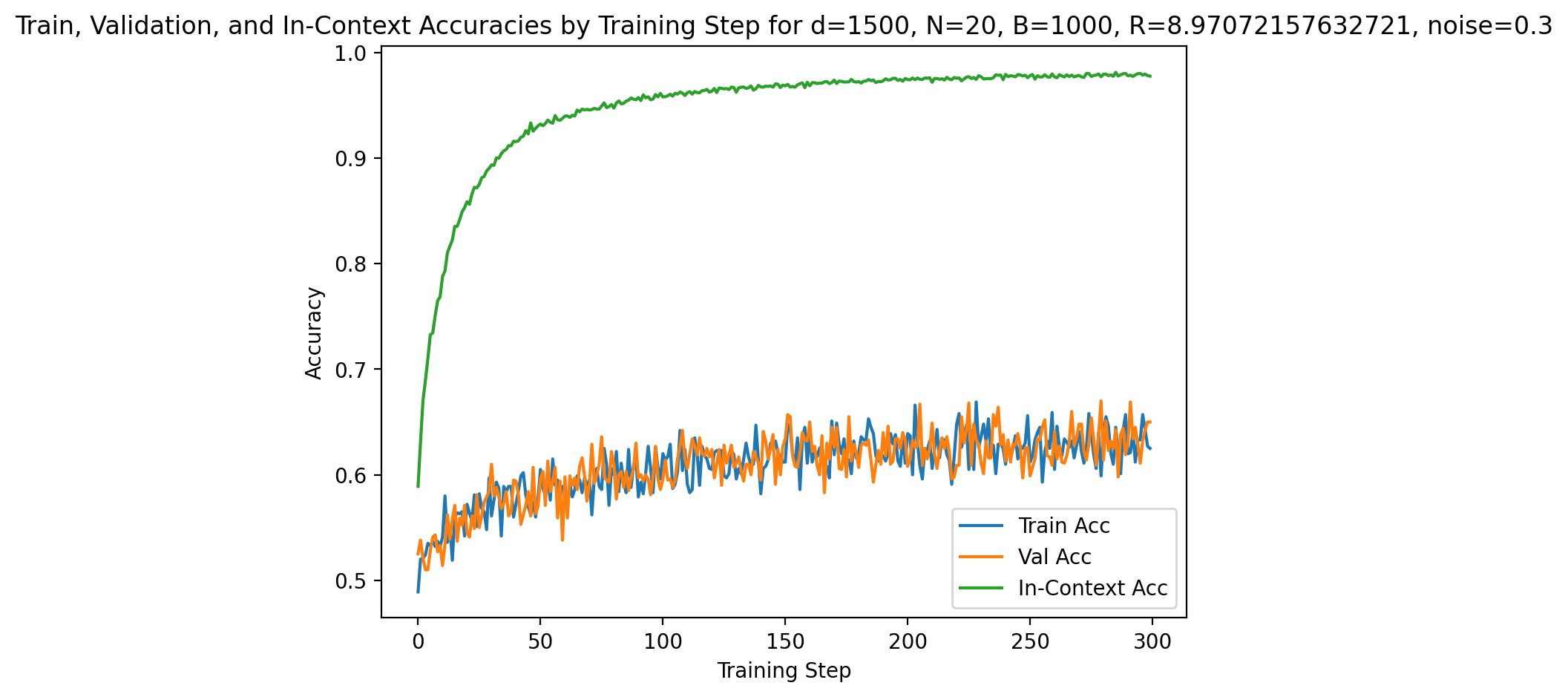}
  \caption{Model Performance (Non-Benign Overfitting Case)}
  \label{fig:high_dim_1}
\end{figure}
\begin{figure}[H]
  \centering
  \includegraphics[width=1.1\textwidth]{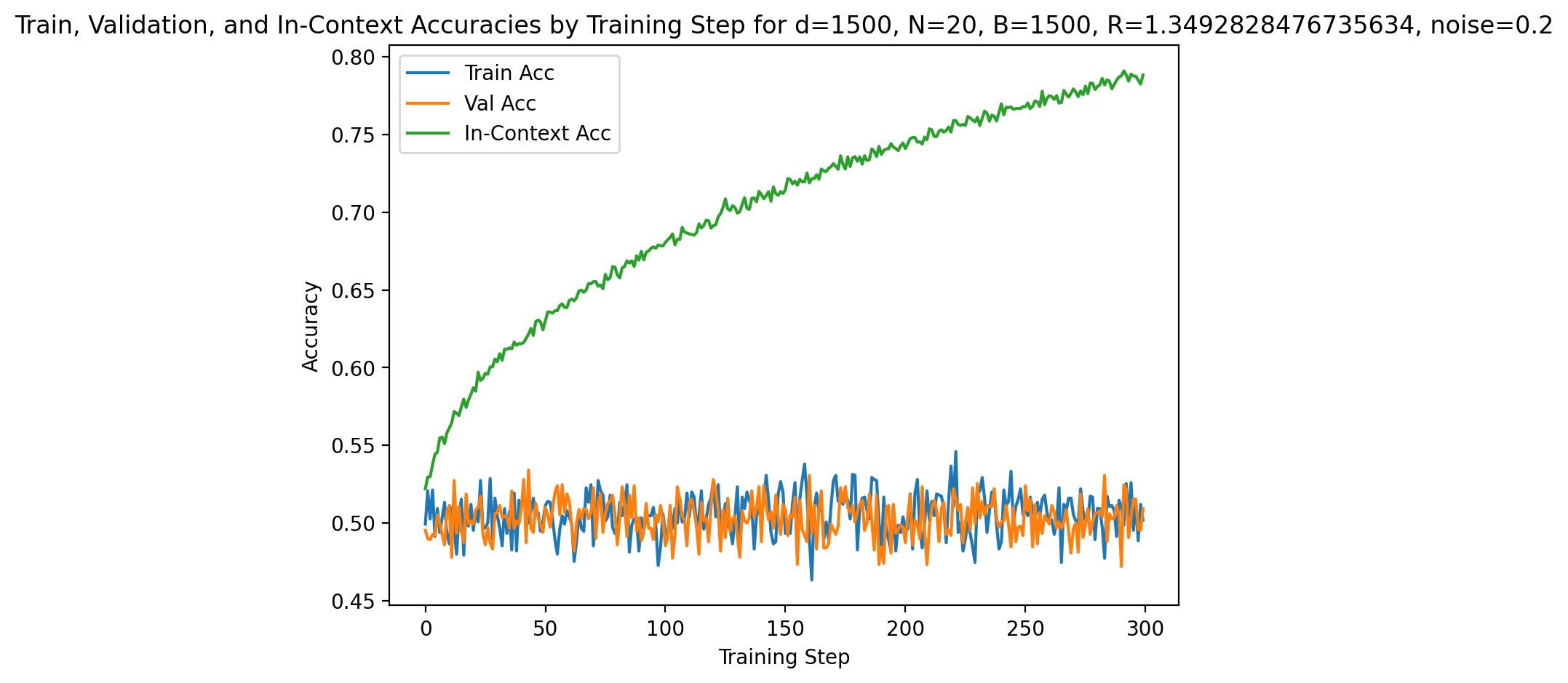}
  \caption{Model Performance (d=1500, N=20, B=1500, R=1.35, noise=0.20)}
  \label{fig:high_dim_1}
\end{figure}
\begin{figure}[H]
  \centering
  \includegraphics[width=1.1\textwidth]{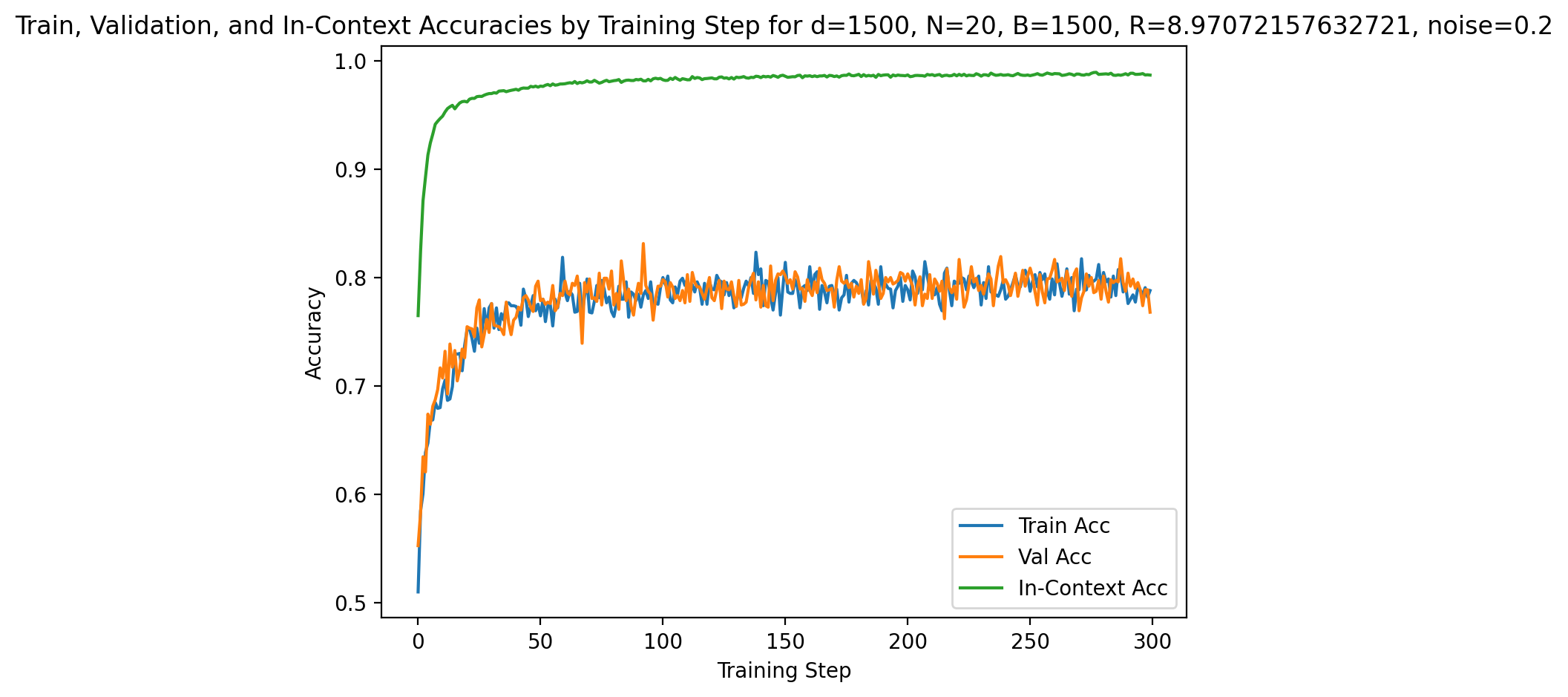}
  \caption{Model Performance (d=1500, N=20, B=1500, R=8.97, noise=0.20)}
  \label{fig:high_dim_1}
\end{figure}
\begin{figure}[H]
  \centering
  \includegraphics[width=1.1\textwidth]{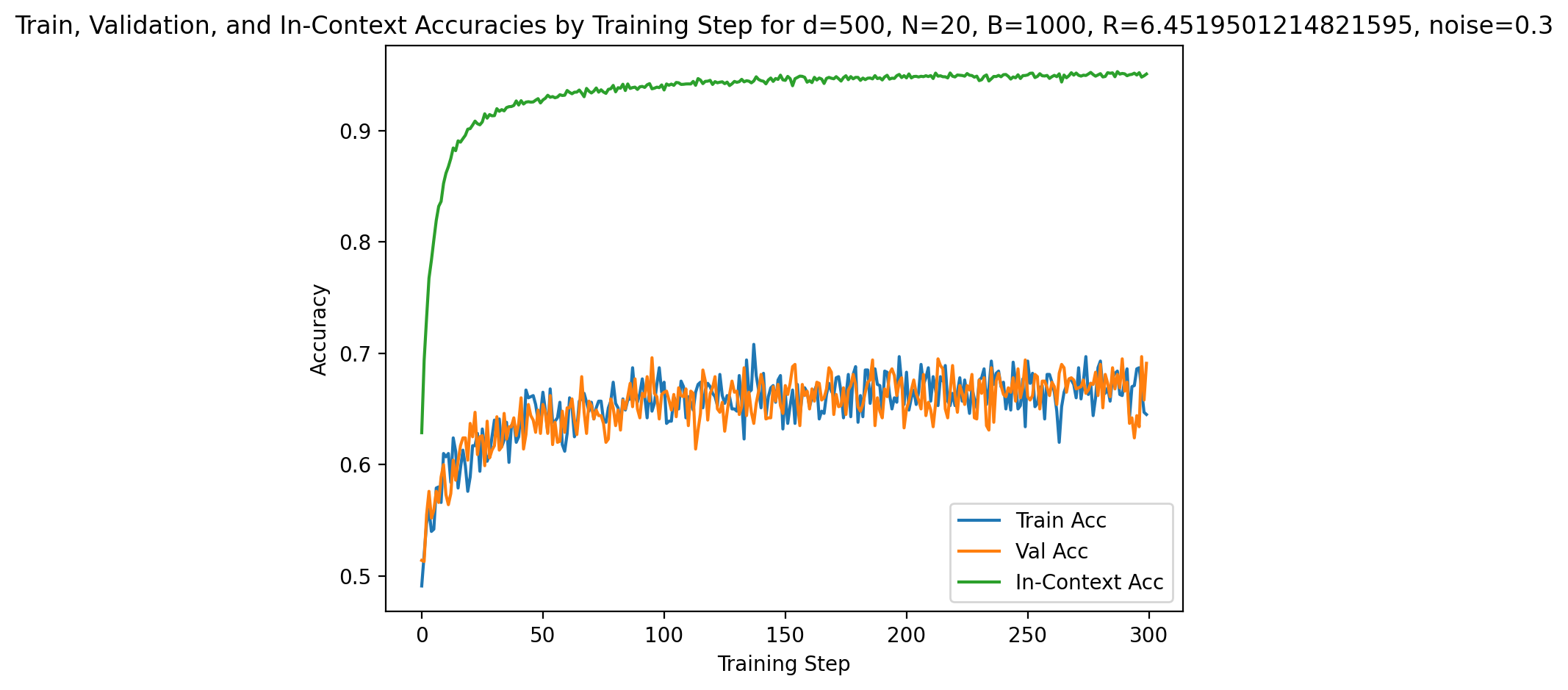}
  \caption{Model Performance (d=500, N=20, B=1000, R=6.45, noise=0.30)}
  \label{fig:high_dim_1}
\end{figure}
\begin{figure}[H]
  \centering
  \includegraphics[width=1.1\textwidth]{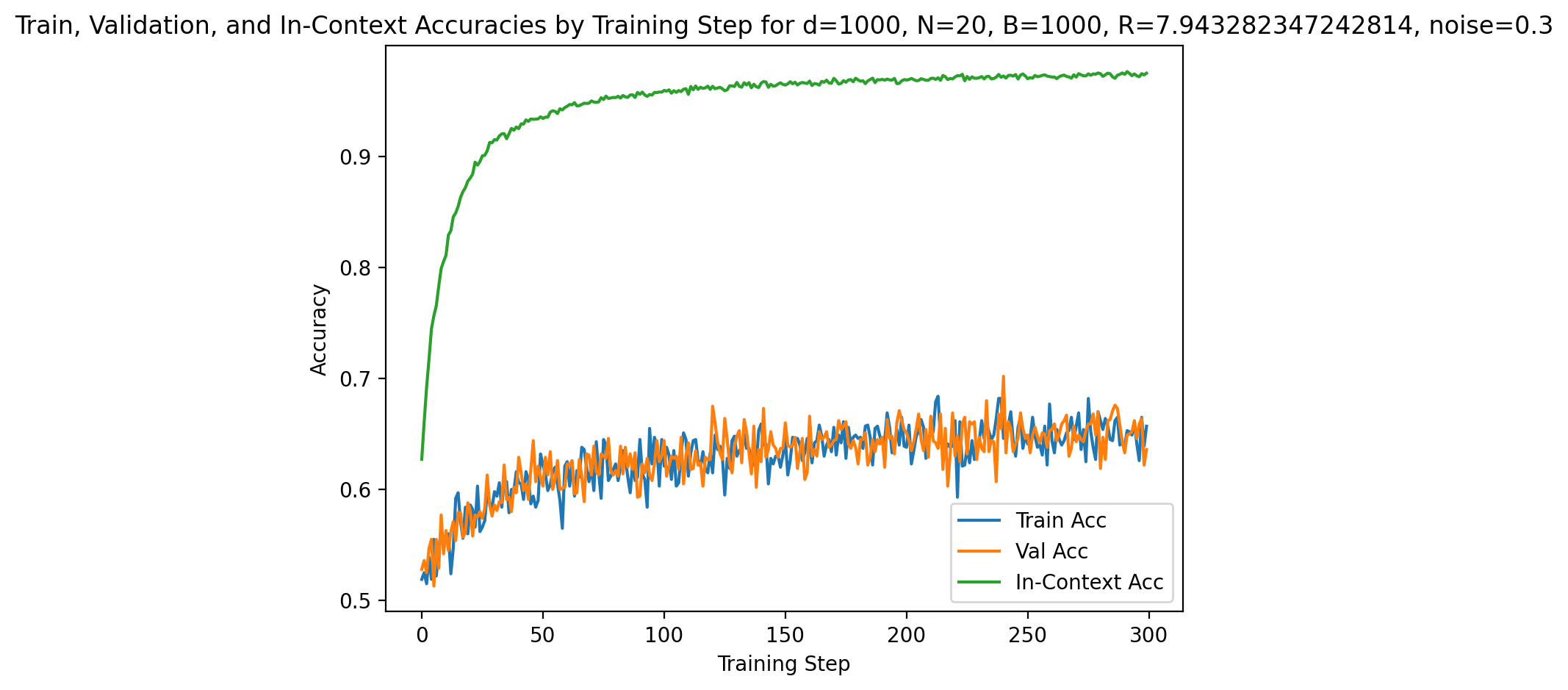}
  \caption{Model Performance (d=1000, N=20, B=1000, R=7.94, noise=0.30)}
  \label{fig:high_dim_1}
\end{figure}

\subsection{Research Question 3}
\begin{figure}[H]
  \centering
  \includegraphics[width=1.1\textwidth]{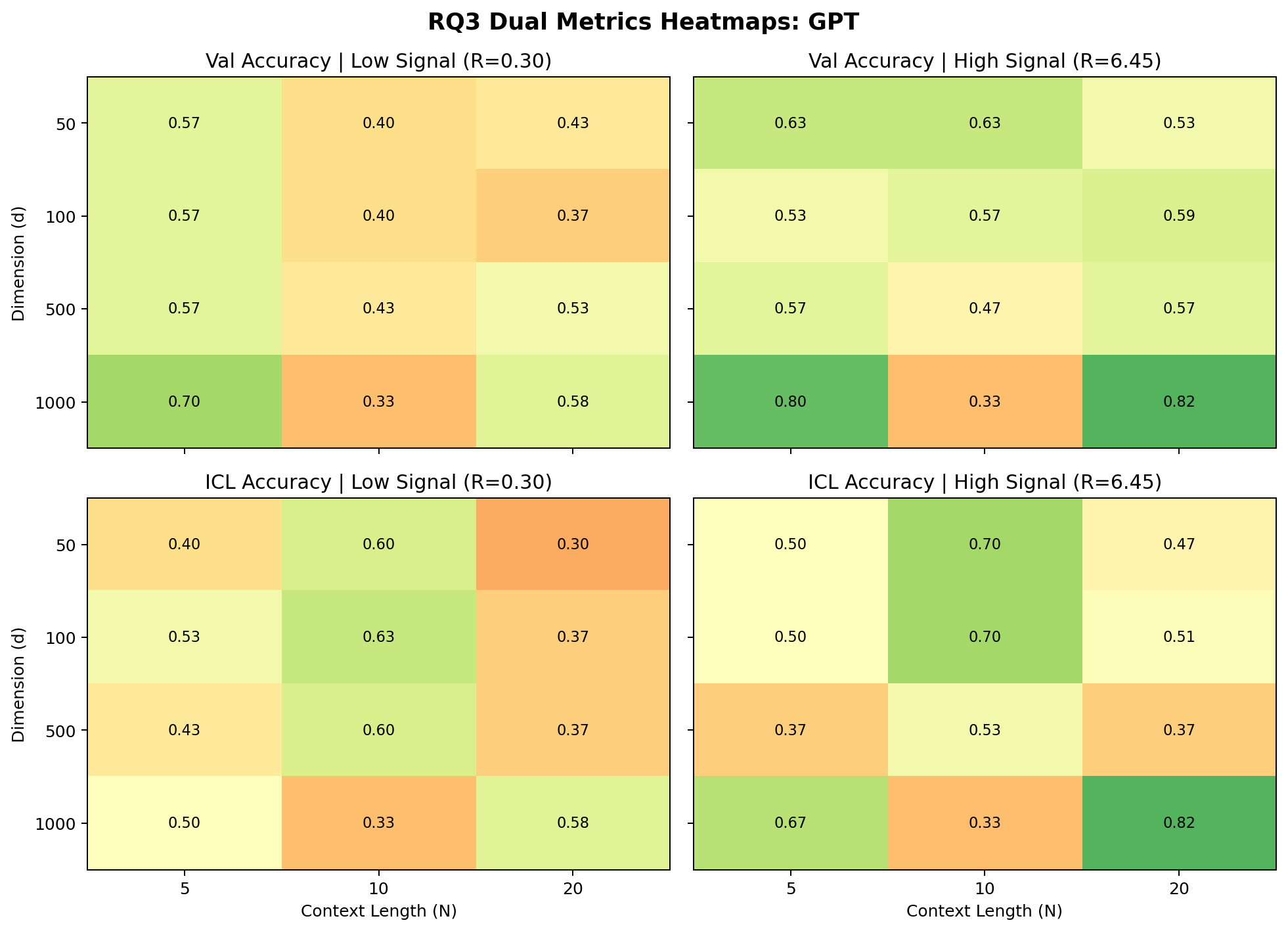}
  \caption{ChatGpt-4-mini results}
  \label{fig:high_dim_1}
\end{figure}

\begin{figure}[H]
  \centering
  \includegraphics[width=1.1\textwidth]{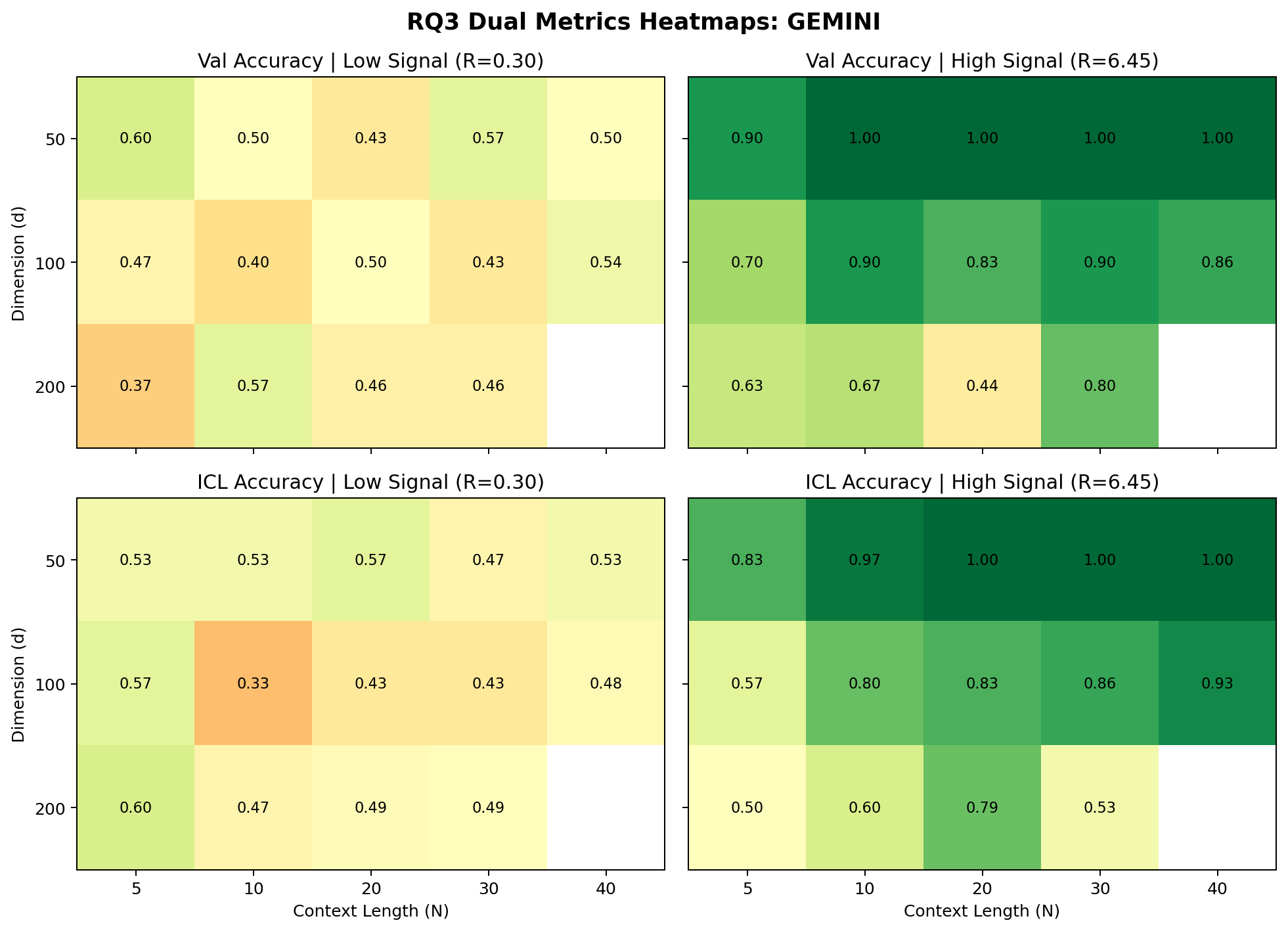}
  \caption{Gemini-2.0-mini results}
  \label{fig:high_dim_1}
\end{figure}

\begin{figure}[H]
  \centering
  \includegraphics[width=1.1\textwidth]{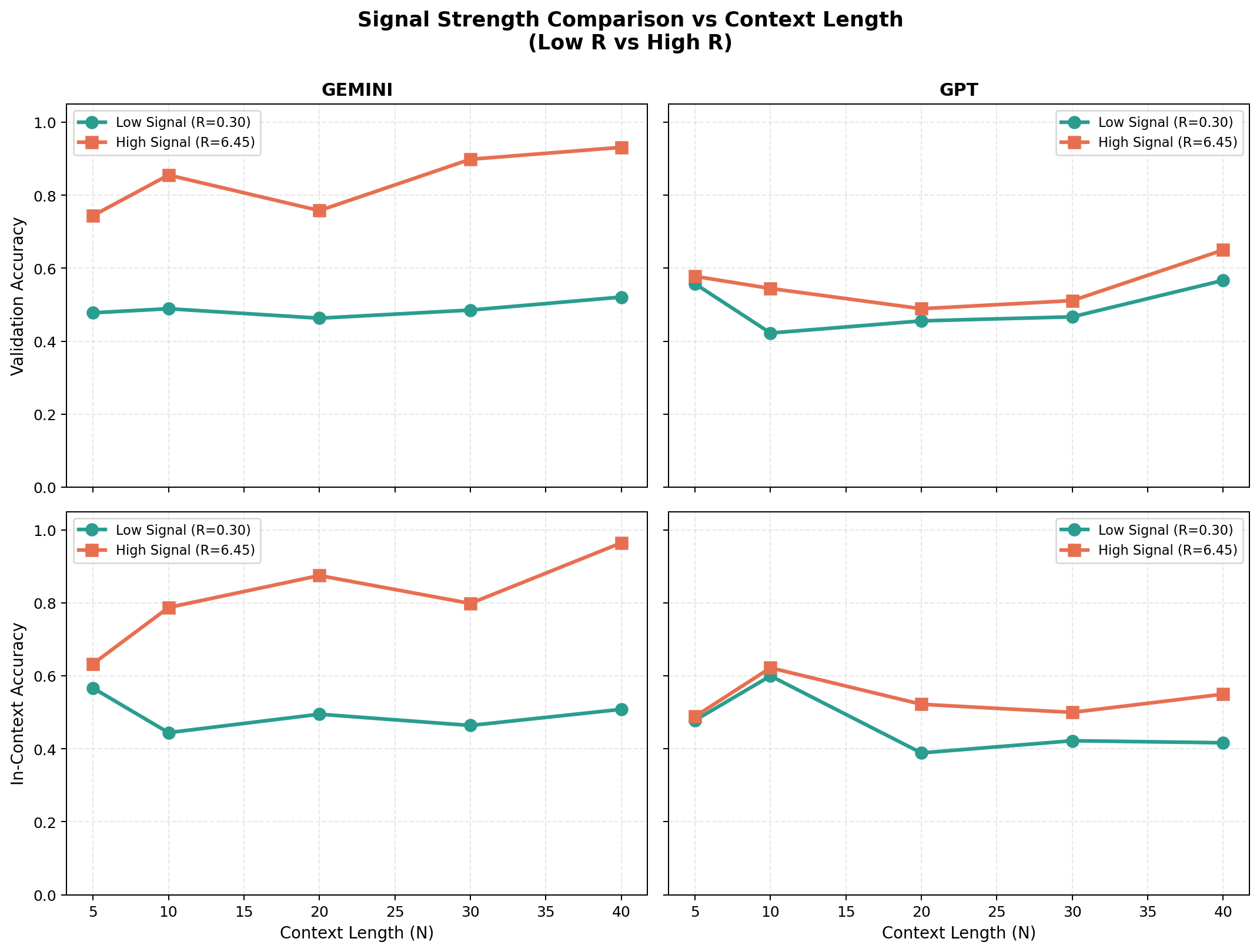}
  \caption{Gemini-2.0-mini results}
  \label{fig:high_dim_1}
\end{figure}

\begin{figure}[H]
  \centering
  \includegraphics[width=1.1\textwidth]{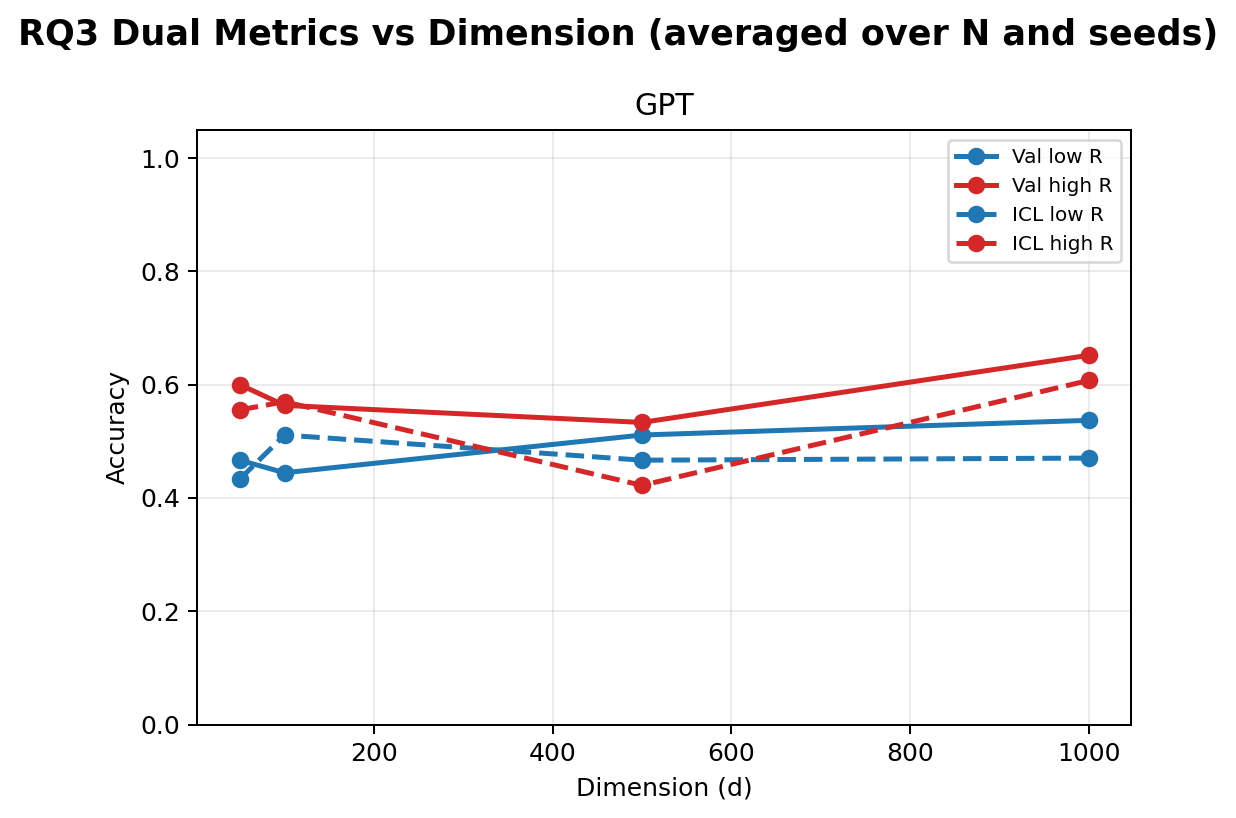}
  \caption{ChatGpt-4-mini results agregated as dimenstion increases}
  \label{fig:high_dim_1}
\end{figure}

\begin{figure}[H]
  \centering
  \includegraphics[width=1.1\textwidth]{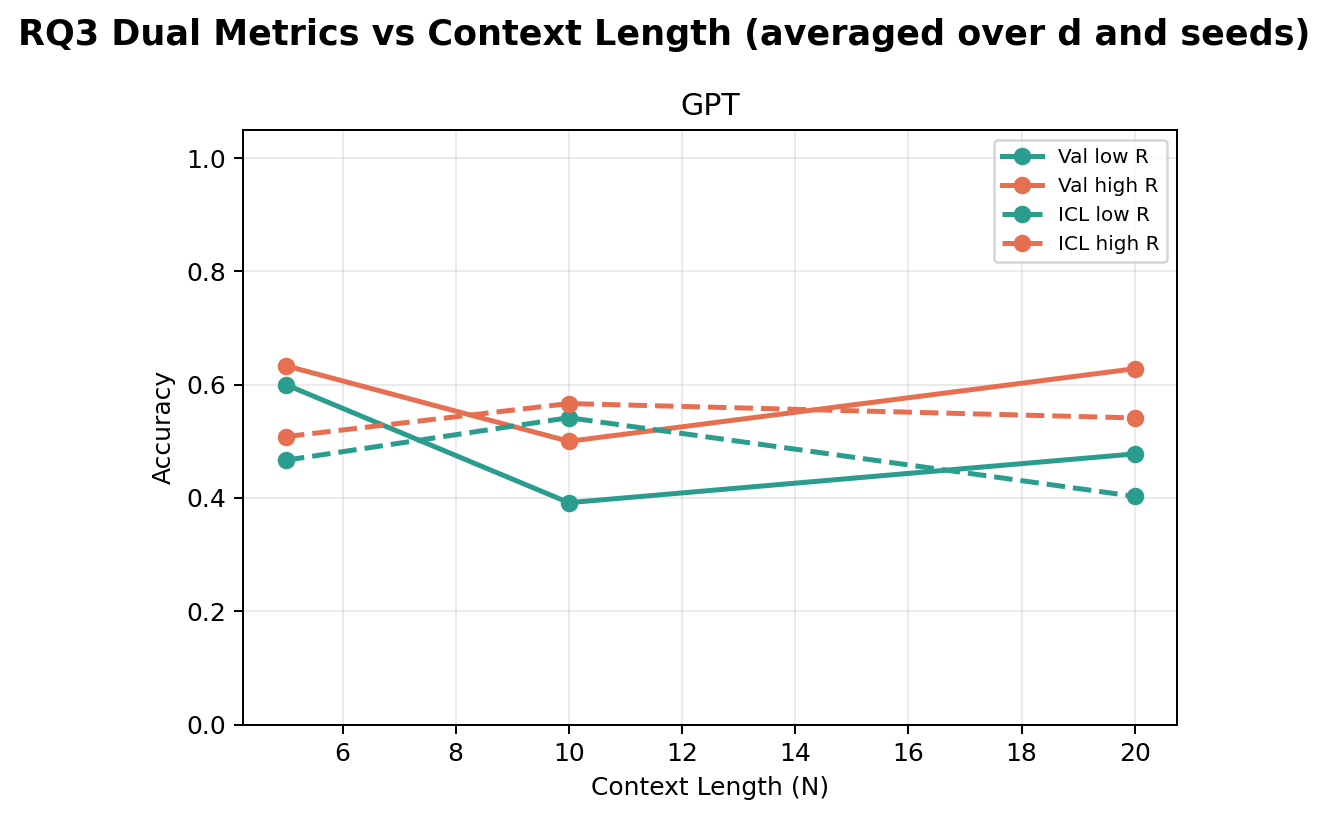}
  \caption{ChatGpt-4-mini validation vs ICL as context length increases}
  \label{fig:high_dim_1}

\end{figure}

\begin{figure}[H]
  \centering
  \includegraphics[width=1.1\textwidth]{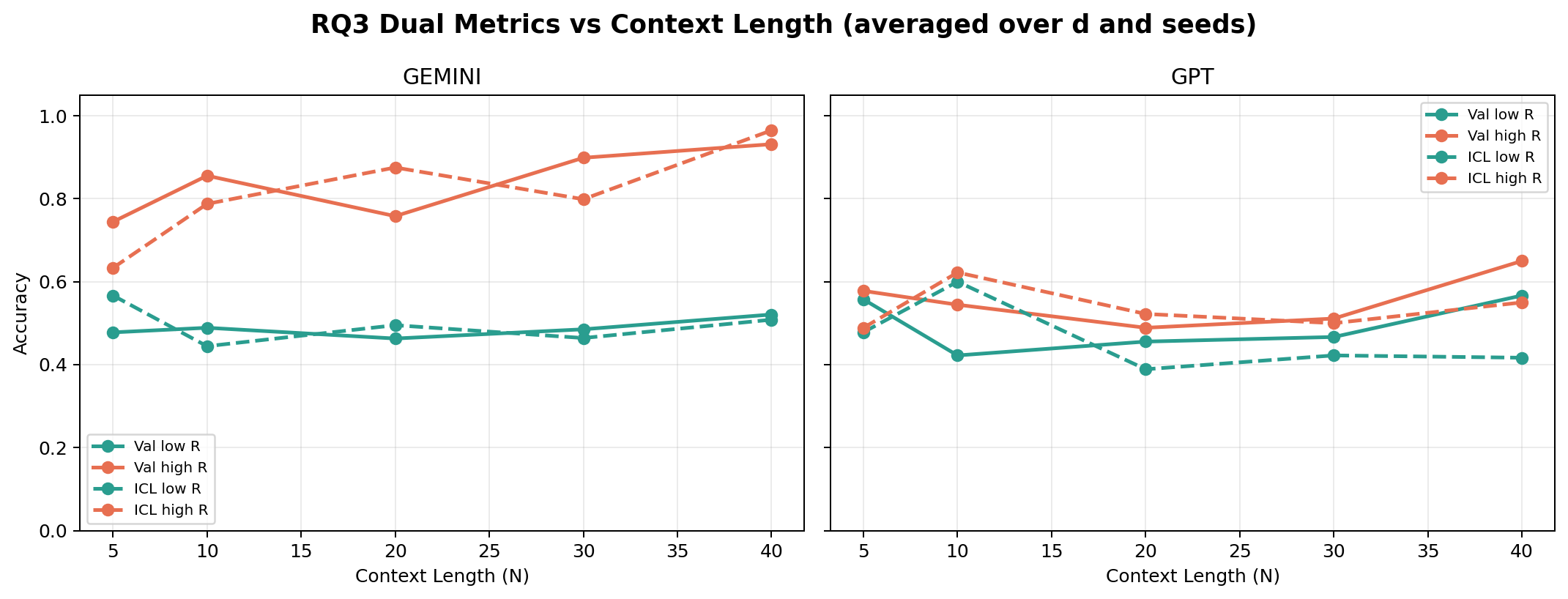}
  \caption{ChatGpt-4-mini vs Gemini 2.0 flash as context examples increase}
  \label{fig:high_dim_1}
\end{figure}
\section{Discussion}
\subsection{Research Question 1}
\paragraph {Dimension Experiments} Figure 1 shows model performance under a low dimension (d=50). We also set R to be scaled with dimension, so that a higher dimension does not automatically equal an easier task since the signal strength will change as well. When R is constant, we expect an easier task under low dimensions and a harder task under higher dimensions. This figure shows that in-context accuracy, or accuracy of unseen tasks that fit the decision boundary, starts high and reaches 1.0 very quickly. However, the train and validation (which also measures in-context accuracy, the accuracy of completely unseen tasks) accuracies hit a bottleneck around 0.95 and do not increase, meaning that under these conditions, the model can classify its own context points but struggles to generalize to query points, or unseen points that did not influence the boundary. Figure 2 shows a similar setup, except with R being set to a constant value. We see that this becomes a much easier task as R=6.45 is a very strong signal strength relative to the 50-dimensional setup, and all accuracies reach 1.0 very quickly. Figure 3 also uses the same constant R=6.45, but increases the dimensions to 1000. Here, we can see more clearly the validation and in-context accuracies reaching 1.0 through the training process, with in-context performance once again reaching that mark faster than the validation accuracy. This performance is justified again as the in-context accuracy tasks are more relevant to the ones the model was trained on versus the more general validation accuracy tasks, but both perform excellently here. Finally, Figure 4 uses the same setup as Figure 3 except setting R to be scaled with dimension once again. We see even better performance in this graph, notably as R ends up being roughly 9.5 after being scaled with the higher dimension, and the signal is even stronger. To conclude, when R is constant, both validation and in-context accuracy are able to reach 1.0 with our Linear Model and classification task despite our dimension value, but a higher dimension will slow convergence because a fixed R becomes weaker in higher-dimensional noise. Under a high dimension and a low R, we will also see performance degrade and not reach the optimal 1.0 because the noise increases as dimension does. Under SNR scaling, performance will always reach 1.0 as the signal strength is large enough to offset the dimension increase.

\paragraph{Sequence Length Experiments} Figure 5 shows model performance under a low sequence length (n=5) while keeping dimension (d) and number of tasks (b) fixed and R constant. We see in-context accuracy start at 0.99 and stay there, while validation accuracy starts much lower at 0.75 but is able to reach the 0.99 mark by step 20. This shows us that having only 5 context examples does not bring down the prediction accuracy of the context points, but it does affect the generalization power of the model to make the same accurate predictions for the query points. Figure 6 shows a similar setup with the exception of SNR. We see that in this setup, the R value does not affect the performance much. Validation accuracy takes a few more steps to hit 1.0, but this difference is essentially negligible. We conduct another test keeping the dimension and number of tasks fixed once again, but this time increasing n to 80 and testing with constant R in Figure 7 and SNR in Figure 8. Both these graphs are also similar to one another and the n=5 setup, with the main difference being that validation accuracy starts higher at roughly 0.98 from step 0. From this, we conclude that sequence length has an effect on the starting position of validation accuracy, representing the model's performance on unseen tasks that were not used to influence the decision boundary. These accuracies tend to start lower when n is also lower, but this does not seem to have a significant effect on the rate at which the models learn. On the other hand, the in-context accuracy, representing the model's performance on unseen tasks but were used to influence the decision boundary, is largely unaffected by sequence length.

\paragraph{Batch Size Experiments} Figure 9 shows model performance under a low batch size, or number of tasks (b=50) while keeping dimension (d) and sequence length (n) fixed and R constant. We see that in-context accuracy starts at 0.75 and validation accuracy starts lower at 0.56, but both are able to reach 1.0 by step 25. This convergence is gradual and unaffected by the noise that causes the training data here to oscillate between 0.6 and 1.0. Figure 10 has a similar setup except with SNR, and similarly to the sequence length experiments, we see both in-context and validation accuracies starting near the same point and reaching 1.0 by step 25. From this, we can conclude that the fixed R vs SNR does not seem to impact the model performance. Extending this to a larger batch size yields different results. Figure 11 shows a similar setup with a constant R and b=2000. With the number of tasks this high, we see that in-context and validation accuracies start extremely high, at 0.998 and 0.98 roughly and converge to 1.0 very quickly. Figure 12 uses a batch size of 1000 as a middle ground, and sees in-context and validation accuracies start a little lower at 0.995 and 0.95. From these graphs, we can see that the batch size does have an effect on the starting context and query performance of the model, but in all cases we converge to 1.0 very quickly. This effect is similar to that of the sequence length discussed above, as once again we see that the higher value leads to an increase in base in-context accuracies and thus quicker model convergence. The R value being fixed or SNR did not seem to affect the performance when only batch size was changed.

\subsection{Research Question 2}

\paragraph {Noise in Context Labels} We find that when noise is applied only to context labels and leaving the query labels clean, the model is able to exhibit some form of benign overfitting across nearly every parameter configuration. Figure 13 shows a configuration with noise = 0.2, meaning that 20 percent of the context labels have a flipped sign. We see that despite the model memorizing these noisy context labels, we still reach 1.0 in-context and validation accuracies. Figure 15 shows a setup in which in-context and validation accuracies reach 1.0, but train accuracy does not get there until 200 steps in. This is identical to the setup in Figure 14, except Figure 14 only uses 100 dimensions versus Figure 15's 1000. We see that as a result, Figure 14 sees all of its accuracies reach 1.0 very quickly compared to Figure 15. Higher dimensions give the model more difficulty in predictions, but does not seem to stop benign overfitting from occurring, unless the dimension range exceeds the R which gives the classes separability. We see that when noise is applied only to the context labels, benign overfitting takes longer in higher dimensions but still occurs. Figure 16 also yields an interesting conclusion when compared to Figure 13, with the configuration difference between the two being double the noise present in Figure 16. We see that train accuracy is actually the one most affected, and it never reaches 1.0. This is strong benign overfitting, since 40 percent of the context labels are flipped but yet the model is still able to memorize the noisy context labels it sees at evaluation and generalize well to clean queries.

\paragraph {Uniform Noise in Context and Query Labels} We find benign overfitting to be more easily observed in configurations where noise is applied more uniformly towards context and query labels. This is seen clearly in figures 19 and 20. Figure 19 showcases a configuration where benign overfitting does not occur. While the noise present is only 0.2, we see that in-context accuracy climbs to 0.75 but the train and validation accuracies stay around 0.5. In a binary classification problem, this is no better than random chance. This is an example of underfitting - the model does not memorize the noisy labels and cannot generalize either. We note the signal strength R as 1.35. Figure 20 shows a similar setup except increasing the signal strength R to 8.97. We see that this one change boosts in-context accuracy to 0.99, while validation accuracy is able to reach its theoretical maximum of 0.8. This graph does display benign overfitting, as the model memorizes noisy labels but is still able to make accurate predictions. Therefore, we can conclude that signal strength, which controls class separation, is critical for benign overfitting and must be strong enough for the model to both memorize noisy labels and maintain generalization, something which was not possible for the configuration in Figure 19. We also see this present in Figure 14, where train accuracy is stuck around 0.95 which is due to the SNR resolving R to 3.0, which is lower than the optimal for this setup. Furthermore, Figure 18 shows a setup with 1500 dimensions, just 500 more than Figure 15 with a similar R. We see that in this case, validation accuracy collapses to 0.63. Another such instance is observed in Figures 21 and 22, with Figure 21 having half the dimensions and almost reaching the theoretical maximum for validation accuracy (0.68 on 0.3 noise). However, Figure 22 uses a slightly higher R but validation accuracy starts degrading more, down to 0.63. We do not see benign overfitting happening in Figure 22, but we do see it under a similar configuration in Figure 17, which has the strong enough R to facilitate this. We can conclude that when noise is applied to query labels also, a higher dimensionality can cause collapse in the model instead of facilitating benign overfitting as it happens when noise is applied only to context labels, assuming that the signal strength is not scaled with dimension.

\subsection{Research Question 3}
\textbf{}
RQ3 dual-metric results indicate a consistent but nuanced separation between query accuracy (validation/generalization behavior) and in-context accuracy (leave-one-out context reconstruction). Across the newer GPT runs, query accuracy is often moderate to high, while ICL accuracy is more variable and frequently lower, suggesting that successful task-level prediction does not always imply strong reconstruction of context labels. In practical terms, the model appears to capture enough task signal to classify held-out query points, but not always enough to reliably recover hidden in-context labels under the leave-one-out protocol. This generalization-vs-memorization asymmetry is one of the strongest patterns in the current dataset and appears across several (d, N, R) configurations rather than being an isolated artifact of a single run.

Signal strength (R) generally helps, but the improvement is not strictly monotonic for every configuration. In lower-dimensional and mid-dimensional settings, moving from low signal (R=0.3) to high signal (R=6.45) tends to stabilize or improve query accuracy, while ICL gains are smaller and less consistent. For example, representative runs show query moving upward while ICL remains flat, and other runs where both metrics improve but query still outpaces ICL. This suggests that stronger class separation provides cleaner decision boundaries for held-out prediction, but does not uniformly solve in-context reconstruction complexity. In short, higher R improves task signal quality, but the two metrics are not interchangeable and should be interpreted separately in reporting.

Dimension (d) and context length (N) effects also appear configuration-dependent rather than globally linear. Some higher-dimensional settings show strong query performance, but do not guarantee similarly strong ICL behavior; conversely, some lower-dimensional settings exhibit respectable ICL with only moderate query gains. Increasing context length can help when the model effectively uses added examples, but it can also introduce noise sensitivity or prompt inefficiency depending on run stability and sample characteristics. This supports a key methodological conclusion: there is no single “best” hyperparameter direction for all metrics, and the behavior of the model is better described as an interaction between d, N, and R rather than a one-variable trend.
\section{Conclusion}
\textbf{}
In conclusion, this empirical study systematically characterizes the in-context learning (ICL) capabilities of transformers on Gaussian-mixture binary classification tasks. Our findings demonstrate that ICL performance relies heavily on the interplay between dimensionality, sequence length, and signal strength. Specifically, while higher dimensions can slow model convergence, scaling the signal-to-noise ratio effectively mitigates this issue, allowing models to consistently achieve optimal accuracy. Furthermore, our investigation into benign overfitting reveals that transformers can successfully memorize noisy context labels while maintaining strong generalization on clean query data, provided the signal strength is sufficiently high to ensure adequate class separation. When extending this analysis to full commercial transformer architectures, we observed a distinct generalization-versus-memorization asymmetry: models can often accurately classify held-out query points even when they struggle to reliably reconstruct hidden in-context labels. Ultimately, these insights provide a comprehensive empirical map of the geometric conditions necessary for successful task inference from context alone , offering a stronger foundation for leveraging ICL mechanisms to reduce the extensive training time and compute power required in modern machine learning.
\section{Appendix}
Project Proposal: \url{https://www.overleaf.com/read/chhcqqncpnyd#00fa76}

\section{Contributions}
\textbf{Rushil:} Abstract, Introduction, Methods 2.1, Code for Methods 2.1, Results 2.1 and 2.2, Discussion 2.1 and 2.2, Website\\
\textbf{Sebastian:} Methods 2.2, Code for Methods 2.2, Poster\\
\textbf{Leonardo:} Methods 2.3, Code for Methods 2.3, Website\\

\makereference

\bibliography{reference}
\bibliographystyle{style/dsc180bibstyle}

\end{document}